\documentclass{article}
\usepackage[preprint]{main}

\usepackage[utf8]{inputenc} 
\usepackage[T1]{fontenc}    
\usepackage{hyperref}       
\usepackage{url}            
\usepackage{booktabs}       
\usepackage{amsfonts}       
\usepackage{nicefrac}       
\usepackage{microtype}      
\usepackage{xcolor}         
\usepackage{subcaption} 

\usepackage{threeparttable}
\usepackage{multirow}
\usepackage{pifont}
\usepackage{makecell}
\usepackage{tcolorbox}
\usepackage{todonotes}
\usepackage{amsmath} 
\usepackage{array}
\usepackage{wrapfig}
\usepackage{enumitem}

\definecolor{mygreen}{rgb}{0.0, 0.7, 0.0}
\definecolor{myred}{rgb}{0.8, 0.0, 0.0} 
\definecolor{mynote}{rgb}{0.0, 0.7, 0.7}

\newcommand{\bdmx}{{\bf x}}

\newcommand{\SBase}{{\bf S}_\text{base}}
\newcommand{\SSyn}{{\bf S}_\text{syn}^{+}}
\newcommand{\SReal}{{\bf S}_\text{real}^{+}}

\newcommand{\nbase}{n_\text{base}}
\newcommand{\nsyn}{n_{\text{syn+}}}
\newcommand{\nreal}{n_\text{real+}}
\newcommand{\etal}{et al.}

\title{Exploring the Equivalence of Closed-Set Generative and Real Data Augmentation in Image Classification}

\author{
Haowen Wang \thanks{Equal contribution. Work was done when H. Wang and G. Zhang were visiting students at UC, San Diego; they are currently with Tsinghua University.}\\
University of California, San Diego \\
\And   
Guowei Zhang \footnotemark[1]\\
University of California, San Diego \\
\And
Xiang Zhang  \\
University of California, San Diego \\
\And  
Zeyuan Chen \\
University of California, San Diego \\
\And
Haiyang Xu \\
University of California, San Diego \\
\And 
Dou Hoon Kwark \\
University of Illinois at Urbana-Champaign
\And
Zhuowen Tu \\
University of California, San Diego \\
}

\begin{document}

\maketitle

\begin{abstract}
  In this paper, we address a key scientific problem in machine learning: Given a training set for an image classification task, {\bf can we train a generative model on this dataset to enhance the classification performance?} (i.e., closed-set generative data augmentation). We start by exploring the distinctions and similarities between real images and closed-set synthetic images generated by advanced generative models. Through extensive experiments, we offer systematic insights into the effective use of closed-set synthetic data for augmentation. Notably, we empirically determine the equivalent scale of synthetic images needed for augmentation. In addition, we also show quantitative equivalence between the real data augmentation and open-set generative augmentation (generative models trained using data beyond the given training set). While it aligns with the common intuition that real images are generally preferred, our empirical formulation also offers a guideline to quantify the increased scale of synthetic data augmentation required to achieve comparable image classification performance. Our results on natural and medical image datasets further illustrate how this effect varies with the baseline training set size and the amount of synthetic data incorporated.
\end{abstract}

\section{Introduction}
\label{sec:intro}

We consider the image classification task in predicting labeling $y$ for a given input $\bdmx$. The \textit{analysis-by-synthesis} methodology \citep{yuille2006vision} has once been considered as one of the guiding principles for making a variety of inferences~\citep{cootes1995active,tu2002image,fergus2003object}. 
The analysis-by-synthesis principle would expect having powerful generative $p(\bdmx|y)$ 
can substantially improve the inference of $p(y|\bdmx)$.  

There is an explosive development with increasing level of maturity in image generation, including generative adversarial learning \citep{tu2007learning,goodfellow2014generative,karras2018progressive}, variational autoencoder (VAE) \citep{kingma2013auto}, and diffusion models \citep{sohl2015deep,ho2020ddpm,rombach2022high,ramesh2021zero}. 
With the increasing representation power and photo-realism of generative modeling, especially diffusion-based models \cite{rombach2022high}, attempts exist to perform generative data augmentation for image classification by augmenting the training set with synthesized images \cite{azizi2023synthetic, fan2024scaling,he2023synthetic,yin2023ttida}. However, previous approaches aim at developing methods and/or showing improved image classification performances using advanced generative models. These methods \cite{azizi2023synthetic, fan2024scaling,he2023synthetic} therefore adopted image models that have been trained/pre-trained with a large amount of external data (i.e., open-set generative data augmentation), in addition to maximizing the assistance from language and vision-language models \cite{yin2023ttida}.

\begin{figure}[!t]
\begin{center}
\includegraphics[width=\linewidth]{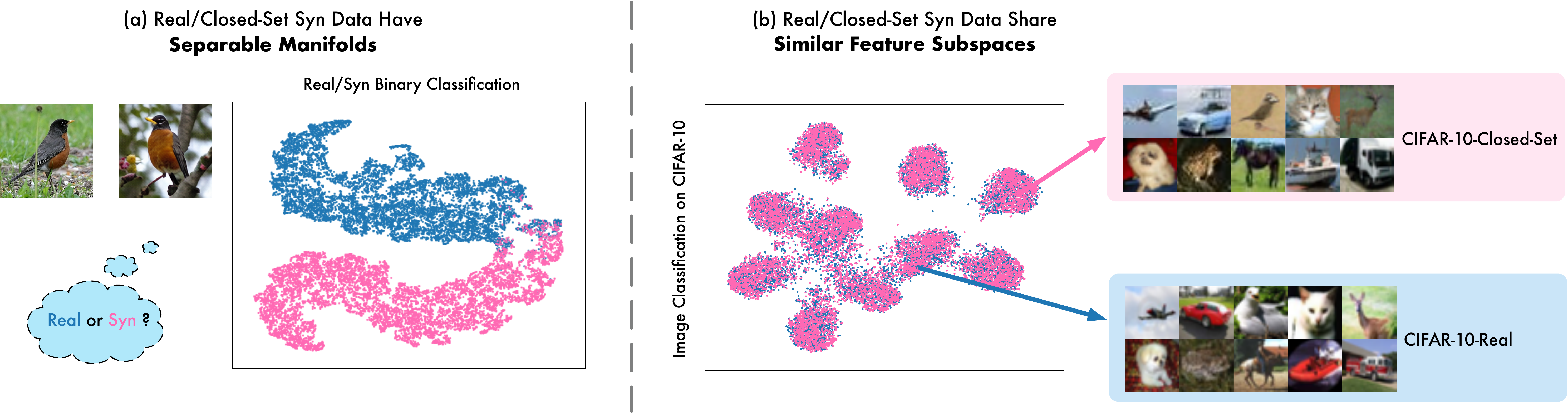}

\vspace{-5pt}
\caption{\small \textbf{The differences and equivalences between real and closed-set synthetic data}: (a) \textbf{Manifold Distinctions}: The manifolds of the real and closed-set synthetic data in subspaces learned by a binary real-syn domain classifier, emphasizing their significant domain gap. (b) \textbf{Feature Subspace Overlap}: Using a standard image classifier trained on CIFAR-10 real images as a feature extractor, the real and closed-set synthetic data exhibit notable overlap in feature subspaces.}
\label{fig:teaser}
\vspace{-12pt}
\end{center}
\end{figure}

We, instead, try to go to the root of image classification by asking a fundamental image representation question: Given an image classification dataset, {\em can we train a generative model on it from scratch to improve classification performance?} If so, {\em how equivalent is augmenting with synthetic data to augmenting with real data?} Our investigation is carried out by studying generative data augmentation from models trained under a closed-set setting, {\bf without using semantic/textual category information}, which helps generalize our findings to a broad range of image classification problems such as biological, neural, and medical data where the {\bf categories can be denoted as simple as Class A and Class B}. 

Moreover, we aim to provide \textbf{quantifiable guidance} on the size of closed-set generative data augmentation, offering a more direct and fundamental approach to evaluating the effectiveness of synthetic data. Suppose we are given a set of training data $\SBase=\{(\bdmx_i,y_i), i=1,..,\nbase\}$, where $\bdmx_i$ indexes the $ith$ image with its corresponding ground-truth label $y_i$. 
Let $\SSyn = \{(\bdmx'_j, y_j), j=\nbase+1,..,\nbase+\nsyn\}$ be an augmented training set of synthesized images where $\bdmx'_j$ refers to each synthesized image; let $\SReal = \{(\bdmx_j, y_j), j=\nbase+1,..,\nbase+\nreal\}$ be an augmented set of real images. We obtain an empirical equivalence for the generative data augmentation size $|\SSyn|=\nsyn$ w.r.t. the real data augmentation size $|\SReal|=\nreal$ in the form of:
    \begin{equation}
    {\frac{\nsyn}{\nbase} \simeq c_1^{\tau\nbase} \times \left(c_2^{\frac{\nreal}{\nbase}}-1 \right)}.
    \label{eq:esti_overall_form}
    \end{equation}
where \(\tau\) is a given coefficient related to the dataset used, while \(c_1\) and \(c_2\) are the parameters to be fitted.

\begin{figure*}[!htp]
  \centering
\includegraphics[width=0.9\linewidth]
{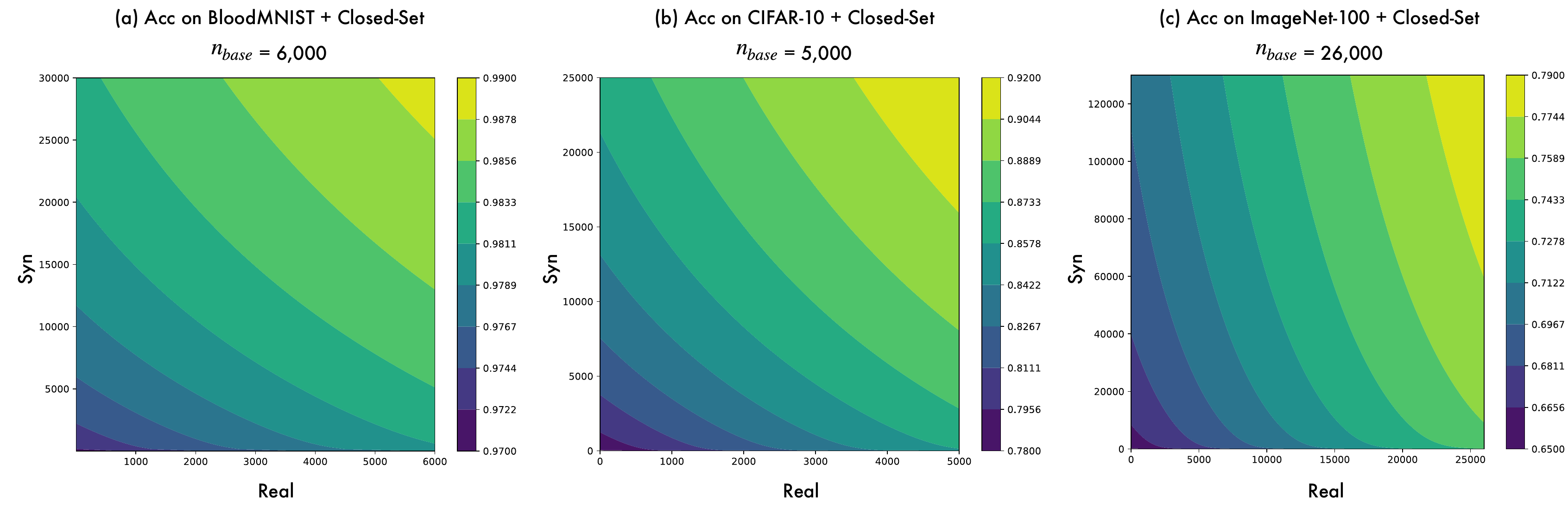}
\vspace{-2pt}
   \caption{{\bf Equivalence curves} w.r.t. the amount of closed-set generative and real data augmentation at fixed $\nbase$ on BloodMNIST, CIFAR-10 and ImageNet-100. Points on each contour line correspond to the same classification accuracy as shown on the color bar.}
   \label{fig:acc_internal_external}
\vspace{-10pt}
\end{figure*}

Fig. \ref{fig:acc_internal_external} shows the corresponding equivalence curves that give rise to fitted forms of Eq. \ref{eq:esti_overall_form} on different datasets.  Further details can be found in Sec. \ref{section: experiments}. It is observed that for closed-set generative data augmentation: 1). \textbf{Adding synthetic data consistently improves classification performance.} However, to achieve the same level of a performance boost, the required \textbf{synthetic data size is always greater than that of the real data}, meaning that having the real data is always more advantageous than using synthetic data; 2). \textbf{The effectiveness of synthetic data scales well when the size of the base training set increases}, implying that it is possible for closed-set generative data augmentation to improve the performance even when the basic classification accuracy is already strong; 3). With a fixed base training set, \textbf{an increase in synthetic data size leads to decreased efficiency}, indicating a diminishing marginal effect of synthetic data. Note that the equations we have fitted (Eq. \ref{eq:esti_internal_bloodmnist}, Eq. \ref{eq:esti_internal_cifar}, and Eq. \ref{eq:esti_internal_imagenet}) are
meant to serve as an empirical quantification for high-level guidance.
This novel study allows us to see the quantitative equivalence of using real vs. closed-set generative data augmentation for image classification.

Although the main focus of our study is on closed-set generative data augmentation, this quantitative equivalence can generalize to open-set scenarios, where we present the equations for two different datasets in Eq. \ref{eq:esti_external_cifar} and Eq. \ref{eq:esti_external_imagenet}.

For the sake of clarity, we define \textit{synthetic} data here as images generated by statistical generative models, distinguishing them from `synthetic' data produced by graphics simulation engines \citep{beery2020synthetic}.

\begin{figure*}[t]
  \centering
\includegraphics[width=0.9\linewidth]{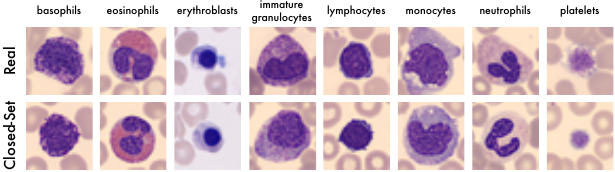}

   \caption{{\bf Visualizations of real and closed-set synthetic data} from the BloodMNIST dataset, where the closed-set synthetic data show great resemblance to the real data.}
   \label{fig:medical_vis}
    \vspace{-15pt}
\end{figure*}

In summary, we present a timely effort to empirically quantify the equivalence between real data augmentation and synthetic data generated in a closed-set manner. To our knowledge, this is the first study of its kind. Importantly, {\bf our major goal is not to demonstrate state-of-the-art numbers on the image classification benchmarks by using modern generative models}, as done in previous works \cite{azizi2023synthetic, fan2024scaling, he2023synthetic, yin2023ttida, yang2022image,singh2024synthetic}. Instead, we aim to {\bf investigate the fundamental problem of image representation} by examining the differences between real images and images created by a generative model trained on a given dataset. In other words, {\bf do generative models carry complementary information beyond that learned by discriminative classifiers} trained on the same dataset? 
Our findings provide practical guidance for general image classification problems, particularly for non-natural images—such as biological, neurological, and medical images—that lack textual category descriptions and cannot be generated by conventional generative models trained on large-scale natural image-text pairs. For example, Fig. \ref{fig:medical_vis_open_set} shows images generated using an open-set generative models \citep{karras2022elucidating} based on textual biology cateogry description, which are significantly deviated from the desirable images displayed in Fig. \ref{fig:medical_vis}.

\section{Related Work}
\label{gen_inst}

\paragraph{Open-Set Generative Data Augmentation}

Recent approaches have explored the use of open-set generative models to generate synthetic data for image classification \citep{sariyildiz2023fake,zhou2023using, bansal2023leaving, hennicke2024mind, jung2024dalda}. Pre-trained text-to-image diffusion models, in particular, have gained prominence as these models can generate high-quality, large-scale curated datasets with just a few textual descriptions.
For instance, He \etal \cite{he2023synthetic} has found that synthetic data generated by GLIDE \cite{nichol2021glide} can readily benefit image classification in data-scarce settings for few-shot learning. Trabucco \etal \cite{trabuccoeffective} proposes a data augmentation method that uses pre-trained text-to-image diffusion models to enhance semantic diversity in images, leading to improved accuracy in few-shot image classification tasks. Azizi \etal \cite{azizi2023synthetic} has demonstrated that fine-tuning Imagen \citep{saharia2022photorealistic} using a target dataset can improve classification accuracy. Fan \etal \cite{fan2024scaling} studies the scaling laws of synthetic images generated by text-to-image diffusion models to train image classifiers. The main objective of \cite{yin2023ttida} is to design an effective synthetic data generation pipeline/method using text-to-text and text-to-image engines to improve the classification results for a number of natural image datasets. 
As stated previously, existing methods primarily focus on improving the current image classification benchmarks using generative models trained on open-set and they {\em have not systematically studied the closed-set generative data augmentation problem}, nor deriving {\em quantitative equivalence between real and synthetic data augmentations}.

\paragraph{Closed-Set Data Augmentation}

We define closed-set data augmentation as the method of performing data augmentation using only the information provided by the given dataset. Traditional closed-set data augmentation techniques include familiar operations such as rotation, scaling, flipping, and cropping. In recent years, more sophisticated methods have emerged, such as Mixup \citep{zhang2017mixup}, which generates new training samples by linearly interpolating between two images, and CutMix \citep{yun2019cutmix}, which creates new training data by cutting and swapping regions between two images. These methods rely on heuristic-based predefined augmentation templates and do not utilize generative models. 

In contrast, our study focuses on closed-set {\bf generative} data augmentation (training a generative model from scratch on the classification dataset) and addresses a fundamental question in image classification: To what extent can synthetic data quantitatively serve as an equivalent to real data augmentation? Additionally, we explore how various factors—such as real and synthetic data scale—impact the effectiveness of data augmentation. Our work diverges from prior generative data augmentation research in its {\bf motivation}, {\bf problem setting}, {\bf experimental design}, and {\bf findings}.

\paragraph{Diffusion Models}

Diffusion models \citep{ho2020denoising, song2020denoising, song2020score} have emerged as powerful generative models capable of producing high-quality, photo-realistic images. Comparing with traditional generative adversarial networks \citep{ goodfellow2014generative, tu2007learning}, diffusion models offer comparable or even superior image quality while also providing greater training stability. Specifically, text-to-image (T2I) diffusion models enable flexible language prompts to generate diverse and customized images. Imagen \citep{saharia2022photorealistic}, Stable Diffusion \citep{rombach2022high} and DALL-E \citep{ramesh2021zero} are notable T2I models. Additionally, ControlNet \citep{zhang2023controlnet}, UniControl \citep{qin2023unicontrol}, and OmniControlNet \citep{wang2024omnicontrolnet} demonstrate excellent capabilities in image-conditioned T2I tasks. 

For our study, we use two generative models for closed-set generative data augmentation: EDM \citep{karras2022elucidating}, which offers a well-defined design space that distinctly separates the key design choices of diffusion models during both the sampling and training processes, achieving near SOTA results on CIFAR-10 \citep{krizhevsky2009learning} image generation; and DiT \citep{Peebles2022DiT}, a latent diffusion model that replaces the commonly-used U-Net backbone with a transformer operating on latent patches. DiT reduces computational demands while demonstrating impressive scalability on ImageNet \citep{deng2009imagenet} image generation.

\section{Initial Observations}
In closed-set generative data augmentation, classification models and generative models share the same source of real training data, so any observed distribution gap between real and synthetic data should be attributed to \textbf{inherent limitations in the generative model itself}. To validate the existence of this gap, we employ a straightforward method to highlight the differences between the real and synthetic data distributions.
Specifically, we train a ResNet-110 model \citep{he2016deep} to classify between real and synthetic images generated by an EDM model \citep{karras2022elucidating} on CIFAR-10. Our results show that the binary classification accuracy using high-quality synthetic data exceeds \textbf{98\%}. Ideally, a generative model should produce synthetic data that is statistically indistinguishable from real data, capturing the full complexity and diversity of the dataset. However, these findings suggest that, due to inherent limitations of the generative model, real and synthetic data exist on distinct manifolds, although synthetic data may appear visually convincing. The space of all valid images is immense, and any existing generative models can only cover a small subspace of the sampling space, resulting in fundamental differences in the statistics of the image patches between synthetic and real. To further validate the distributional differences, we use t-SNE \citep{van2008visualizing} to visualize the feature vectors of real and synthetic images extracted by the binary classifier, as shown in Fig. \ref{fig:teaser}(a), where the distribution gap is clearly significant. Additional experiments and analyses are provided in Appendix \ref{appendix:fundamentally_different}.

On the other hand, if we train a standard image classifier on CIFAR-10 real images and use it to extract features from real and closed-set synthetic data, we observe significant overlaps (as shown in Fig. \ref{fig:teaser}(b)), suggesting that despite distributional disparities, synthetic data closely resembles real data for image classifiers. This observation raises important questions: Can closed-set data improve classification performance? If so, how is closed-set generative data augmentation equivalent to real data augmentation? These questions form the central focus of our study.

\section{Analysis of Closed-Set Generative Data Augmentation}
\label{section: experiments}
In this section, we outline the experiments conducted to explore the equivalence of generative and real data augmentation, accompanied by a comprehensive set of conclusions. We first introduce the overall paradigm of our experiments, followed by a detailed examination of closed-set generative data augmentation in the contexts of non-natural and natural image classification. In the natural image setting, we further perform a comparative analysis with open-set scenarios.
\begin{table*}[t]
  \vskip 0.05in
  \centering
  \begin{threeparttable}
  \begin{small}
  \renewcommand{\multirowsetup}{\centering}
  \renewcommand{\arraystretch}{1.1}
  \setlength{\tabcolsep}{3.5pt}
  \caption{A summary table of experimental settings in this study, with detailed descriptions provided in later sections respectively.}
  \label{tab:summary}
  \begin{tabular}{c|c|c|c|c|c}
    \toprule
   Natural or & \multirow{2}{*}{{Real Dataset}} & 
   {Generative} & Generative Model/ &
    \multirow{2}{*}{{Base Training Set Size}} & \multirow{2}{*}{{
    Classifier}}\\
      
     Non-natural? & {} & {Augmentation}  & {Synthetic Dataset} & {} & {}\\
    \toprule
    
    \multirow{4}{*}{{Natural}} & \multirow{2}{*}{{CIFAR-10}} & Closed-Set & EDM, class-cond. & \multirow{2}{*}{{100\% / 50\% / 10\% / 1\%}} & \multirow{2}{*}{{ResNet-110}}\\
    
    & & Open-Set & CIFAKE & & \\
    \cmidrule{2-6}
    & \multirow{2}{*}{{ImageNet-100}}& Closed-Set & DiT, class-cond.& \multirow{2}{*}{{100\% / 20\% / 5\%}} & ResNet-50 \& \\
    
    & & Open-Set & SD3, captions& & ViT-B/32\\
    
    \midrule
    Non-Natural & BloodMNIST & Closed-Set & EDM, class-cond.& 100\% / 50\% / 10\% & ResNet-110\\
    \bottomrule
  \end{tabular}
    \end{small}
    \vspace{-10pt}
  \end{threeparttable}
  
\end{table*}

\subsection{Overall Experimental Paradigm}\label{section:cifar}

For each classification dataset (e.g., CIFAR-10), we conduct experiments under different base training set sizes to simulate varying amounts of available images in practical classification problems. In the case of real data augmentation, additional unused images from the original dataset are incorporated into the base training set. For generative data augmentation, we either train a generative model from scratch on the base training set (closed-set scenario) or use pre-trained text-to-image generative models (open-set scenario). Notably, in the closed-set setting, even for the same classification dataset (e.g., CIFAR-10), we retrain the generative model separately for each base training set size. Synthetic images, generated at different ratios, are mixed with real images to train the classifier backbone. Unless specified otherwise, all evaluations are conducted on the whole validation set of the original dataset. The specific settings for all experiments are summarized in Tab. \ref{tab:summary}.

For convenience, we will refer to the base training sets of different sizes, from largest to smallest, as \{dataset name\}-Large (full scale), \{dataset name\}-Mid, \{dataset name\}-Small and \{dataset name\}-Tiny.

\subsection{Non-Natural Image Classification}

For non-natural image classification, we utilize the BloodMNIST dataset \citep{yang2023medmnist}, which contains microscope images of 8 types of normal blood cells. For more details, please refer to the original paper \citep{yang2023medmnist}. We used the 64x64 resolution version and resized it to 32x32 for classification. For closed-set generative data augmentation, we employ a diffusion model with advanced training and sampling techniques introduced in EDM \citep{karras2022elucidating}, training it exclusively on the subsets of the resized dataset to generate synthetic data. The classifier is evaluated on the test set of BloodMNIST \citep{yang2023medmnist}. The visualizations of the real and closed-set synthetic data are shown in Fig. \ref{fig:medical_vis}. Qualitatively, the closed-set synthetic data shows strong resemblance to the real data.

\paragraph{Empirical Equivalence}
We explore how much additional closed-set synthetic data $\nsyn$ is equivalent to a given amount of additional real data $\nreal$ under a fixed $\nbase$. Specifically, we first fit the relationship between classification accuracy and the amount of real and closed-set synthetic data.  The resulting contour plot for BloodMNIST-Mid ($\nbase=6,000$) is shown in Fig. \ref{fig:acc_internal_external}(a). Using the fitted accuracy function, we then determine the synthetic data amount $\nsyn$ that achieves the same accuracy as $\nreal$ added real samples, leading to the derivation of the formula in Eq. \ref{eq:esti_internal_bloodmnist}. Please refer to Appendix \ref{sec:appendix_equations} for more details on the fitting method and the data used. 

\begin{equation}
    \underbrace{\frac{\nsyn}{\nbase} \simeq 0.72^\frac{\nbase}{1,200} \times \left(3.88^{\frac{\nreal}{\nbase}}-1 \right)}_{\text{BloodMNIST: closed-set}} \;
    \label{eq:esti_internal_bloodmnist}
    \end{equation}
\vspace{-5pt}
\begin{figure}[t]
  \centering
  \begin{minipage}[b]{0.48\linewidth}
    \centering
    \includegraphics[width=\linewidth]{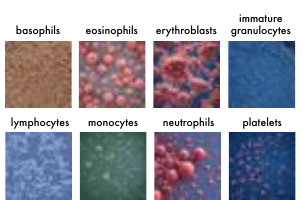}
    \caption{{\bf Visualizations of open-set synthetic data.} Using biology category textual prompts for Stable Diffusion 3 \citep{esser2024scaling} produces images that deviate significantly from the desired content.}
    \vspace{-2pt}
    \label{fig:medical_vis_open_set}
  \end{minipage}
  \hfill
  \begin{minipage}[b]{0.48\linewidth}
    \centering
    \includegraphics[width=\linewidth]{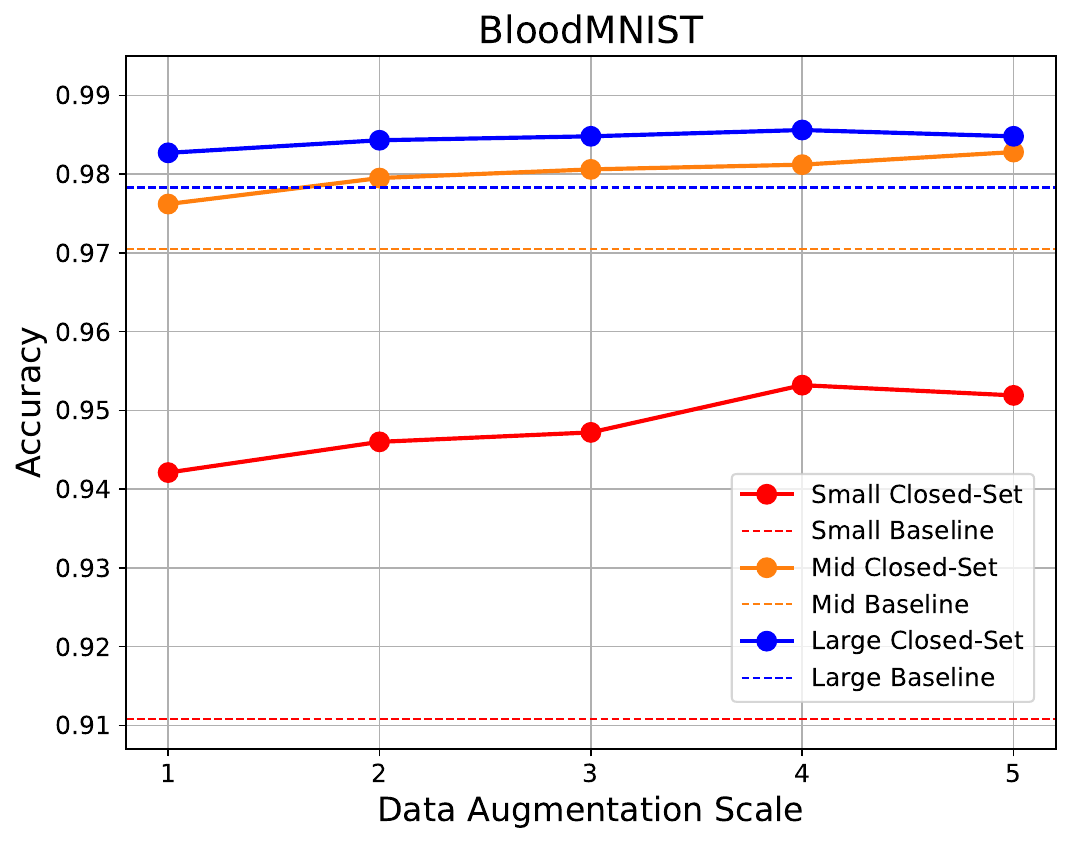}
    \caption{{\bf Accuracy curves.} Synthetic data augmentation scale on BloodMNIST-Small, Mid, and Large.}
    \label{fig:acc_medical}
  \end{minipage}
\vspace{-12pt}
\end{figure}

\paragraph{Main Results}
The experimental results are shown in Fig. \ref{fig:acc_medical}. Closed-set generative data augmentation consistently enhances performance across different values of $\nbase$. Notably, on BloodMNIST-Large, adding 4x synthetic data increases accuracy from 97.83\% to \textbf{98.56\%}, highlighting its effectiveness for non-natural image classification.

Additionally, to demonstrate that open-set generative data augmentation fails to accurately generate blood cell images, we also prompted Stable Diffusion 3 \citep{esser2024scaling} with the template $p_c=$ ``A highly detailed, realistic microscopic image of \textit{c}'', where $c$ represents the blood cell type. 
However, for Stable Diffusion 3 \citep{esser2024scaling}, terms like \textit{basophils} and \textit{eosinophils} carry little textual information, functioning similarly to generic labels such as \textit{Class A} and \textit{Class B}, which limits the model's ability to generate biologically meaningful images.
The visualization of the open-set synthetic data is shown in Fig. \ref{fig:medical_vis_open_set}.

\subsection{Natural Image Classification}

\subsubsection{CIFAR-10 Classification}
To conduct our experiments on natural image classification, we use the full CIFAR-10 training set (CIFAR-10-Large) along with three subsets: CIFAR-10-Mid (50\%), CIFAR-10-Small (10\%), and CIFAR-10-Tiny (1\%). In the closed-set setting, similar to BloodMNIST, we trained EDM models \citep{karras2022elucidating} from scratch on the aforementioned CIFAR-10 subsets for generative data augmentation.

\begin{figure*}[t]
  \centering
  \vspace{-5pt}
\includegraphics[width=\linewidth]{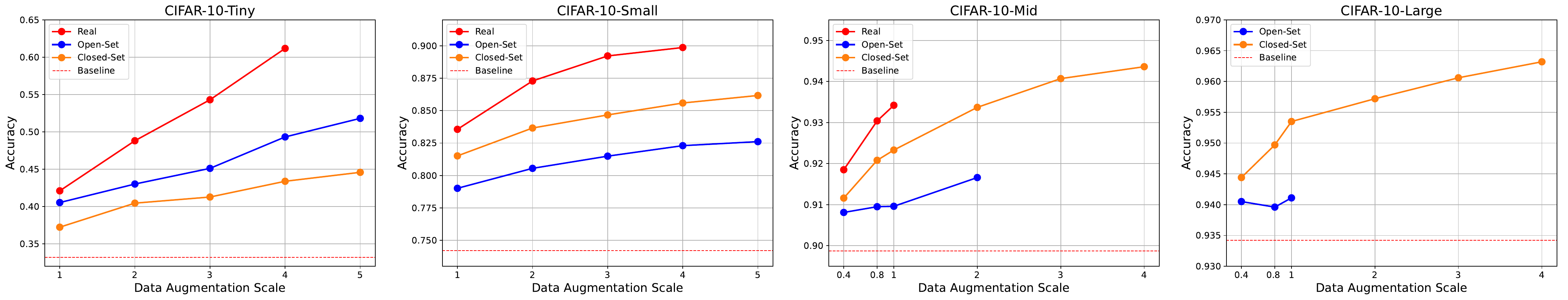}
    
   \caption{{\bf Accuracy curves} w.r.t. the data augmentation scale (real or syn) on CIFAR-10 at different base training set sizes. A data augmentation scale of $k$ means adding extra data to the base training set at a real-to-syn ratio of 1:$k$.}
   \label{fig:acc_cifar}
   \vspace{-10pt}
\end{figure*}

For comparison, we incorporate synthetic data from the CIFAKE dataset \citep{bird2024cifake}, which is generated using Stable Diffusion 1.4 \citep{rombach2022high}, a model pre-trained on a subset of LAION-5B \citep{schuhmann2022laion}. We use a ResNet-110 model as our classification backbone. The results are shown in Fig. \ref{fig:acc_cifar}.
\paragraph{Empirical Equivalence}Similarly, based on the results on CIFAR-10, we analyze the relationship between accuracy and the amount of real and synthetic data at various fixed values of $\nbase$. A contour map is provided in Fig. \ref{fig:acc_internal_external}(b) for CIFAR-10-Small ($\nbase = 5,000$). We then fitted the equivalence equations, as shown in Eq. \ref{eq:esti_internal_cifar} and Eq. \ref{eq:esti_external_cifar}, to roughly assess the effectiveness of synthetic data.
    \begin{equation}
    \underbrace{\frac{\nsyn}{\nbase} \simeq  0.88^\frac{\nbase}{5,000} \times \left(2.53^{\frac{\nreal}{\nbase}}-1 \right)}_{\text{CIFAR-10: closed-set}}
    \label{eq:esti_internal_cifar}
    \end{equation}
    \begin{equation}
    \underbrace{\frac{\nsyn}{\nbase} \simeq 1.67^\frac{\nbase}{5,000} \times \left(2.93^{\frac{\nreal}{\nbase}}-1 \right)}_{\text{CIFAR-10: open-set}}
    \label{eq:esti_external_cifar}
    \end{equation}

\subsubsection{ImageNet-100 Classification}
For closed-set generative augmentation on ImageNet-100, we use the class-conditional DiT-XL/2 model \citep{Peebles2022DiT} that outperforms U-Net-based diffusion models. The original DiT used the pre-trained VAE  \citep{kingma2013auto} from Stable Diffusion \citep{rombach2022high} and fine-tuned it on the target dataset; however, we retrained the VAE to enforce a strict closed-set. The DiT blocks were trained from scratch on the VAE-encoded features. A CFG scale of 4 was used for sampling.

Our open-set data generation protocol uses Stable Diffusion 3 \citep{esser2024scaling} prompted by diverse captions generated according to the method described in \citep{tian2024learning}. The caption templates include $c \rightarrow caption$, $c, bg \rightarrow caption$, and $c, rel \rightarrow caption$. We refer readers to the original paper for further details on this method. Each caption generates five images, and we employ the CLIP-Filter strategy \citep{he2023synthetic} to exclude the bottom 20\% of images based on CLIP zero-shot classification confidence, retaining only the high-quality images. 

\begin{figure*}[t]
    \centering
    \begin{minipage}{0.238\textwidth}
        \centering
        \includegraphics[width=\linewidth]{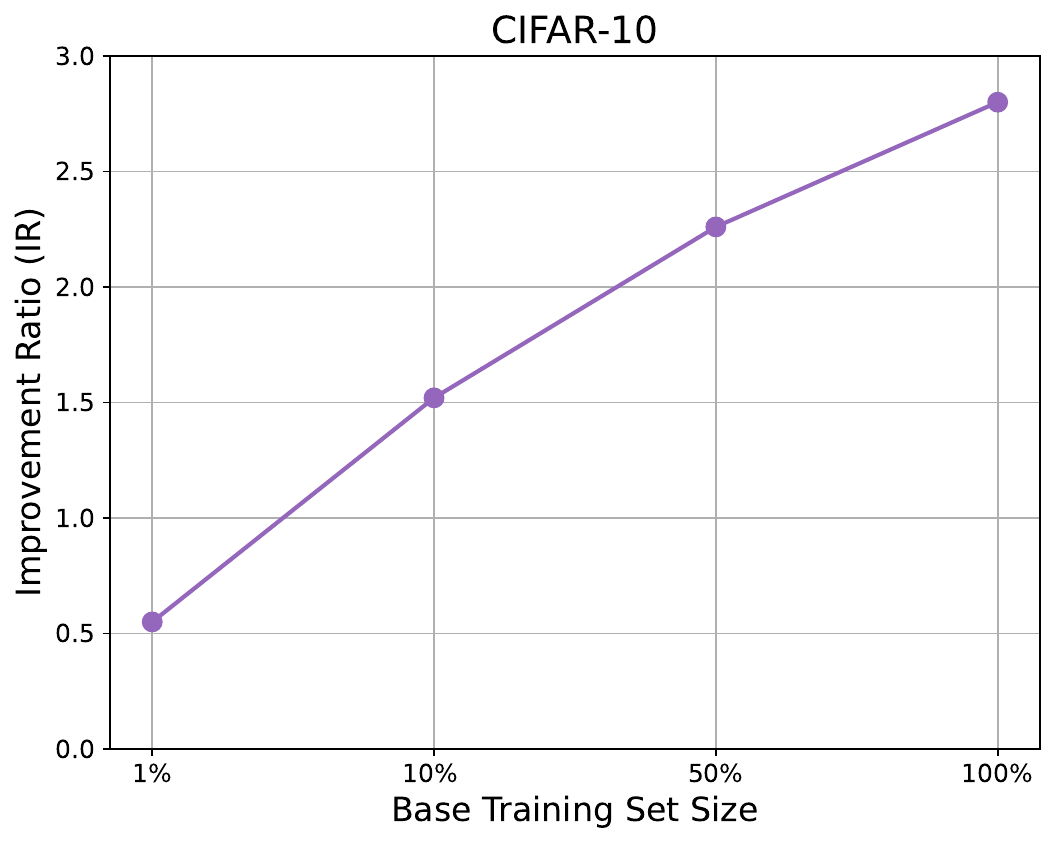}
        
   \caption{{\bf IR} w.r.t. the base training set size on CIFAR-10.}
   \vspace{-10pt}
    \label{fig:ir_cifar}\end{minipage}\hfill
    \begin{minipage}{0.735\textwidth}
        \centering
    \includegraphics[width=\linewidth]{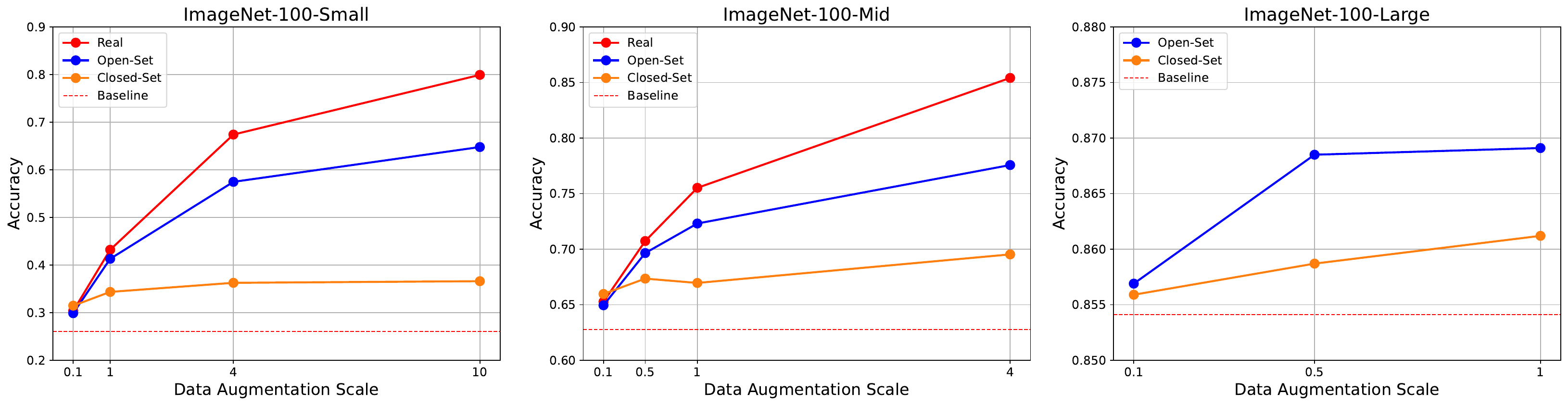}
    
   \caption{{\bf Accuracy curves} w.r.t. the data augmentation scale (real or syn) on ImageNet-100 at different base training set sizes.}
   
\label{fig:acc_imgnet}
    \end{minipage}
    \vspace{-5pt}
\end{figure*}

We use a ResNet-50 model \citep{he2016deep} for all experiments in this section. The accuracy curves are presented in Fig. \ref{fig:acc_imgnet}. Additionally, we conducted generalization experiments using a ViT-B/32 \cite{dosovitskiy2020image} classifier, as detailed in Sec. \ref{sec: ablation}.

\paragraph{Empirical Equivalence}
Similar to the previous experiments, we obtain the closed-set and open-set empirical equivalence equations for ImageNet-100 as below. We provide the contour map for ImageNet-100-Mid ($\nbase=26,000$) in Fig. \ref{fig:acc_internal_external}(c).

    \begin{equation}
    \underbrace{\frac{\nsyn}{\nbase} \simeq 0.94^\frac{\nbase}{13,000} \times \left(37.84^{\frac{\nreal}{\nbase}}-1 \right)}_{\text{ImageNet-100: closed-set}} \;
    \label{eq:esti_internal_imagenet}
    \end{equation}
    \begin{equation}
    \underbrace{\frac{\nsyn}{\nbase} \simeq 1.51^\frac{\nbase}{13,000} \times \left(1.68^{\frac{\nreal}{\nbase}}-1 \right)}_{\text{ImageNet-100: open-set}}
    \vspace{-5pt}
    \label{eq:esti_external_imagenet}
    \end{equation}

\begin{figure*}[t]
  \centering
\includegraphics[width=0.9\linewidth]{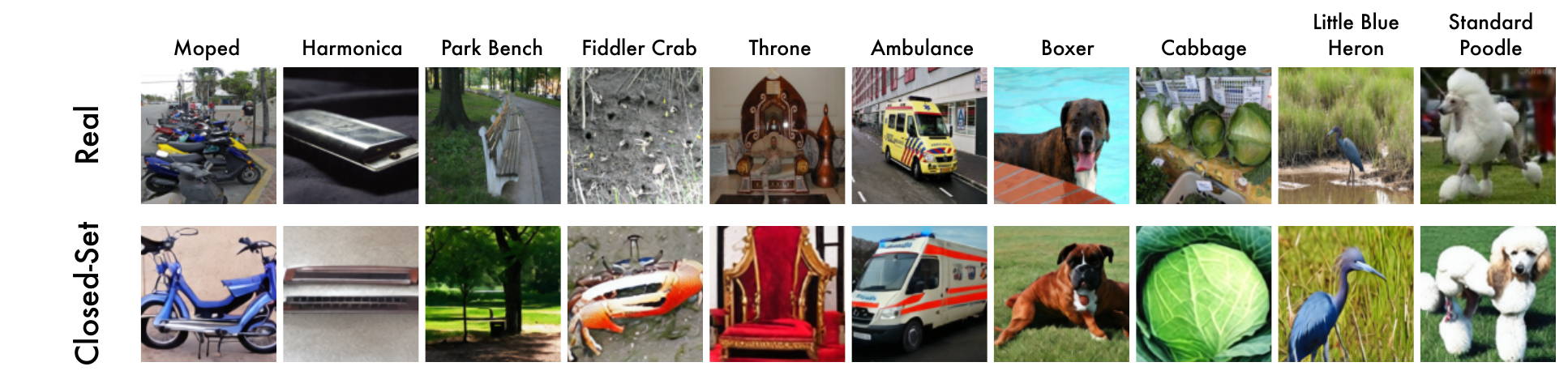}
   \caption{{\bf Visualizations of real and closed-set synthetic data} from the ImageNet-100 dataset.}
   \label{fig:imgnet_100_large_closed_set}
   \vspace{-10pt}
\end{figure*}

\vspace{-10pt}
\subsubsection{Discussion}\label{section:discussion}
In this section, we will summarize the patterns of closed-set generative data augmentation derived from the accuracy curves and equivalence formulas.
\begin{tcolorbox}[colback=blue!3!white, boxrule=1.2pt, colframe=white!10!black, rounded corners,  boxsep=-1pt]
\textit{Closed-set generative data augmentation consistently enhances classifier performance, sometimes even surpassing open-set configurations.}
\end{tcolorbox}

As shown in Fig. \ref{fig:acc_cifar} and Fig. \ref{fig:acc_imgnet}, closed-set generative data augmentation consistently enhances classification performance by a non-trivial margin across CIFAR-10 and ImageNet-100. Surprisingly, although CIFAR-10-Tiny and ImageNet-10-Small only contain less than 100 samples per class, resulting in low-quality generated images (see Appendix \ref{sec:appendix_cifar-10_syn_data} for qualitative results), the improvement achieved is still significant. We hypothesize that this is due to the strong inductive bias of generative models. With a small amount of real data, although the generative model may not produce perfectly accurate images, it can still capture key features, such as the shape of a horse or the texture of a frog’s skin. This capability significantly improves the generalization of discriminative models when data is scarce. 

In the accuracy curves on CIFAR-10 (Fig. \ref{fig:acc_cifar}), we observe that, except for CIFAR-10-Tiny, closed-set generative data augmentation consistently outperforms open-set. This is also reflected in the empirical equivalence functions, where the coefficients $c_1$ and $c_2$ of the closed-set setting are both smaller than those in the open-set setting, indicating that the equivalent amount of synthetic data is generally smaller for the closed-set approach on CIFAR-10.

To explain this observation, we computed the FID \citep{heusel2017gans} and IS \citep{salimans2016improved} metrics for both open-set and closed-set generative data augmentation, as shown in Tab. \ref{tab:cifar-10-quality}. Although the IS values for both are similar, the FID score of the closed-set is significantly lower. This suggests that while both augmentation methods produce high-quality and diverse images, the closed-set data better aligns with the real data distribution, leading to improved classification performance.

However, compared to CIFAR-10, the performance of closed-set generative data augmentation on ImageNet-100 is considerably lower than that of open-set. Beyond factors such as model architecture, resolution, and dataset size, we believe a key reason is that images in CIFAR-10 are generally simpler and object-centric, whereas those in ImageNet-100 exhibit high diversity, making it more challenging for generative models to accurately capture their distribution. This results in a high FID and low IS of the closed-set generative augmentation setting in Tab. \ref{tab:imagenet-100-quality}. Qualitatively, even images generated by the DiT-XL/2 model \citep{Peebles2022DiT} trained on the full ImageNet-100 dataset (ImageNet-100-Large) contain many distortions, as shown in Fig. \ref{fig:imgnet_100_large_closed_set}.

\begin{tcolorbox}[colback=blue!3!white, boxrule=1.2pt, colframe=white!10!black, rounded corners,  boxsep=-1pt]
\textit{Closed-set generative data augmentation exhibits better scalability as the base training set size increases.}
\end{tcolorbox}

\begin{table}[t]
\centering
\begin{minipage}{0.48\linewidth}
  \centering
  \begin{threeparttable}
  \begin{small}
  \renewcommand{\multirowsetup}{\centering}
  \setlength{\tabcolsep}{6pt}
  \caption{Data quality comparison on CIFAR-10. 50,000 synthetic images are used to calculate FID and IS.}
  \label{tab:cifar-10-quality}
  \begin{tabular}{c|c|cc}
    \toprule
   \multirow{2}{*}{{Real Dataset}} & {Generative} &
    \multicolumn{2}{c}{\rotatebox{0}{{Quality}}} \\
    \cline{3-4}
     & {Augmentation} & {FID$\downarrow$}  & {IS$\uparrow$}\\
    \toprule
    \multirow{2}{*}{{CIFAR-10}} & Closed-Set & {{\bf 8.33}} & {{\bf 6.23}}\\
    & {Open-Set} & {27.15}  & {6.14}\\
    \bottomrule
  \end{tabular}
  \end{small}
  \end{threeparttable}
\end{minipage}
\hfill
\begin{minipage}{0.48\linewidth}
  \centering
  \begin{threeparttable}
  \begin{small}
  \renewcommand{\multirowsetup}{\centering}
  \setlength{\tabcolsep}{6pt}
  \caption{Data quality comparison on ImageNet-100. 130,000 synthetic images are used to calculate FID and IS.}
  \label{tab:imagenet-100-quality}
  \begin{tabular}{c|c|cc}
    \toprule
   \multirow{2}{*}{{Real Dataset}} & {Generative} &
    \multicolumn{2}{c}{\rotatebox{0}{{Quality}}} \\
    \cline{3-4}
     & {Augmentation} & {FID$\downarrow$}  & {IS$\uparrow$}\\
    \toprule
    \multirow{2}{*}{{ImageNet-100}} & Closed-Set & {{27.78}} & {{15.11}}\\
    & {Open-Set} & {\bf 17.02}  & {\bf 31.59}\\
    \bottomrule
  \end{tabular}
  \end{small}
  \end{threeparttable}
\end{minipage}
\vspace{-10pt}
\end{table}

When $\frac{\nreal}{\nbase}$ is fixed, a larger \( c_1 \) indicates that the ratio \( \frac{\nsyn}{\nbase} \) goes up faster with increasing \( \nbase \). Comparing the formulas for closed-set and open-set generative data augmentation, we find that on both CIFAR-10 and ImageNet-100, the \( c_1 \) parameter of closed-set is smaller than that of open-set. This suggests that as the base training set grows, closed-set generative augmentation exhibits better scalability.

Such behavior is particularly noticeable on CIFAR-10. To effectively illustrate this, we examine the ratio of accuracy improvements when augmenting the base training set with closed-set and open-set synthetic data at a real-to-syn ratio of 1:1. We refer to this improvement ratio as $\mathrm{IR}$, which can be mathematically expressed as:

\vspace{-8pt}
\begin{equation}
    \mathrm{IR}(n_\text{base}) = \frac{\Delta \mathrm{Acc}_{\text{closed-set,1:1}}(n_\text{base})}{\Delta \mathrm{Acc}_{\text{open-set,1:1}}(n_\text{base})}.
\end{equation}

The variation of $\mathrm{IR}$ with $\nbase$
is shown in Fig. \ref{fig:ir_cifar}. As observed, on CIFAR-10-Tiny (1\%), closed-set generative data augmentation underperforms compared to open-set. However, as the base training set expands, the closed-set approach surpasses the open-set, with the improvement ratio continuing to rise, reaching \textbf{2.80} at an augmentation scale of 4. This indicates that closed-set synthetic data augmentation demonstrates strong scalability. We attribute this to the fact that generative models trained from scratch are able to model the image distribution better with more training data available. As a result, closed-set generative augmentation may even become more effective as the base training set size increases.

\begin{tcolorbox}[colback=blue!3!white, boxrule=1.2pt, colframe=white!10!black, rounded corners,  boxsep=-1pt]
\textit{With a fixed base training set, the performance gains from closed-set generative data augmentation diminish rapidly as the augmentation scale increases.}
\end{tcolorbox}

The $c_2$ parameters of Eq. \ref{eq:esti_internal_cifar} and Eq. \ref{eq:esti_internal_imagenet} are both greater than 1.00, which indicates that when $\nbase$ is fixed, the ratio $\frac{\nsyn}{\nbase}$ goes up exponentially with $\frac{\nreal}{\nbase}$. This suggests that as additional closed-set synthetic data is incorporated, its effectiveness gradually saturates.

This trend is most evident in Fig. \ref{fig:acc_imgnet}, where, for a fixed $\nbase$, the growth of the closed-set curve remains extremely slow. Even on CIFAR-10, where closed-set generative augmentation performs relatively well, the gap between it and real data augmentation still continues to widen as the scale of data augmentation increases, as shown in Fig. \ref{fig:acc_cifar}.

We believe one reason for this phenomenon is that the closed-set generative model fails to capture the rich patterns present in real data, resulting in more homogeneous synthetic images. To verify this, we adopt the \textit{diversity} metric proposed in \citep{boutin2022diversity}. This metric extracts features from a ViT classifier \citep{wightman2019timm} pre-trained on ImageNet, computes the standard deviation within the feature space for each class, and then averages the scores across all classes.

To ensure that the evaluated data is not included in the ViT’s training set, we use the validation set of ImageNet-100 for measurement. For synthetic data, we randomly sample a dataset of the same size generated by a DiT model \citep{Peebles2022DiT} trained on ImageNet-100-Large. The results show that the diversity score for real data is 2.09, whereas for synthetic data, it is only 1.17. This suggests that when a large amount of closed-set synthetic data is already present, the high similarity within its internal distribution prevents it from providing additional useful information to the classifier, leading to diminishing improvements in classification performance.

Notably, the $c_2$ parameters in Eq. \ref{eq:esti_external_cifar} and Eq. \ref{eq:esti_external_imagenet} are also greater than 1.00, indicating that the open-set configuration follows the same pattern. This aligns with the conclusions of \citep{he2023synthetic}.

\subsubsection{Generalization Studies}
\label{sec: ablation}
To demonstrate that the conclusions in the above discussion are not limited to ResNet-50, we replaced the classifier architecture with ViT-B/32 \citep{dosovitskiy2020image} and conducted closed-set and open-set generative data augmentation experiments on ImageNet-100. The fitted formulas are shown in Eq. \ref{eq:esti_internal_imagenet_vit} and Eq. \ref{eq:esti_external_imagenet_vit} below. As can be seen, even with the ViT architecture, the $c_1$ and $c_2$ parameters in the equations still reflect patterns consistent with the previous section. The accuracy curves with respect to the base training set size and data augmentation scale are provided in Fig. \ref{fig:appendix_vit}.

\begin{equation}
    \underbrace{\frac{\nsyn}{\nbase} \simeq 0.97^\frac{\nbase}{13,000} \times \left(38.45^{\frac{\nreal}{\nbase}}-1 \right)}_{\text{ImageNet-100: closed-set (ViT)}} \;
    \label{eq:esti_internal_imagenet_vit}
    \end{equation}
    \begin{equation}
    \underbrace{\frac{\nsyn}{\nbase} \simeq 1.48^\frac{\nbase}{13,000} \times \left(2.16^{\frac{\nreal}{\nbase}}-1 \right)}_{\text{ImageNet-100: open-set (ViT)}}
    \vspace{-5pt}
    \label{eq:esti_external_imagenet_vit}
    \end{equation}
    
\section{Conclusion}
\label{section: conclusion}
In this paper, we present an empirical study that systematically explores the equivalence of generative and real image augmentation using diffusion models trained from scratch on the classification dataset. We provide a quantifiable guideline for utilizing synthetic images to serve as equivalent to real images across both closed-set and open-set generative data augmentation situations. 
 
However, the equations we have derived are empirical without theoretical foundations, and due to computational constraints, the datasets used in our study are relatively small. Nevertheless, we hope that our findings will provide valuable insights for future research on utilizing synthetic data for a wide range of image classification tasks beyond easily accessible natural images.

\paragraph{Acknowledgment} This work is supported by NSF Award IIS-2127544 and NSF Award IIS-2433768.

    \small
    \bibliographystyle{plain}
    \bibliography{main}

\begin{thebibliography}{10}

\bibitem{azizi2023synthetic}
Shekoofeh Azizi, Simon Kornblith, Chitwan Saharia, Mohammad Norouzi, and
  David~J Fleet.
\newblock Synthetic data from diffusion models improves imagenet
  classification.
\newblock {\em arXiv preprint arXiv:2304.08466}, 2023.

\bibitem{bansal2023leaving}
Hritik Bansal and Aditya Grover.
\newblock Leaving reality to imagination: Robust classification via generated
  datasets.
\newblock {\em arXiv preprint arXiv:2302.02503}, 2023.

\bibitem{beery2020synthetic}
Sara Beery, Yang Liu, Dan Morris, Jim Piavis, Ashish Kapoor, Neel Joshi, Markus
  Meister, and Pietro Perona.
\newblock Synthetic examples improve generalization for rare classes.
\newblock In {\em WACV}, pages 863--873, 2020.

\bibitem{bird2024cifake}
Jordan~J Bird and Ahmad Lotfi.
\newblock Cifake: Image classification and explainable identification of
  ai-generated synthetic images.
\newblock {\em IEEE Access}, 2024.

\bibitem{boutin2022diversity}
Victor Boutin, Lakshya Singhal, Xavier Thomas, and Thomas Serre.
\newblock Diversity vs. recognizability: Human-like generalization in one-shot
  generative models.
\newblock {\em arXiv preprint arXiv:2205.10370}, 2022.

\bibitem{cootes1995active}
Timothy~F Cootes, Christopher~J Taylor, David~H Cooper, and Jim Graham.
\newblock Active shape models-their training and application.
\newblock {\em Computer vision and image understanding}, 61(1):38--59, 1995.

\bibitem{deng2009imagenet}
Jia Deng, Wei Dong, Richard Socher, Li-Jia Li, Kai Li, and Li~Fei-Fei.
\newblock Imagenet: A large-scale hierarchical image database.
\newblock In {\em CVPR}, 2009.

\bibitem{5206848}
Jia Deng, Wei Dong, Richard Socher, Li-Jia Li, Kai Li, and Li~Fei-Fei.
\newblock Imagenet: A large-scale hierarchical image database.
\newblock In {\em 2009 IEEE Conference on Computer Vision and Pattern
  Recognition}, 2009.

\bibitem{devlin-etal-2019-bert}
Jacob Devlin, Ming-Wei Chang, Kenton Lee, and Kristina Toutanova.
\newblock {BERT}: Pre-training of deep bidirectional transformers for language
  understanding.
\newblock In {\em Proceedings of the 2019 Conference of the North {A}merican
  Chapter of the Association for Computational Linguistics: Human Language
  Technologies, Volume 1 (Long and Short Papers)}, 2019.

\bibitem{dosovitskiy2020image}
Alexey Dosovitskiy, Lucas Beyer, Alexander Kolesnikov, Dirk Weissenborn,
  Xiaohua Zhai, Thomas Unterthiner, Mostafa Dehghani, Matthias Minderer, Georg
  Heigold, Sylvain Gelly, et~al.
\newblock An image is worth 16x16 words: Transformers for image recognition at
  scale.
\newblock {\em arXiv preprint arXiv:2010.11929}, 2020.

\bibitem{esser2024scaling}
Patrick Esser, Sumith Kulal, Andreas Blattmann, Rahim Entezari, Jonas
  M{\"u}ller, Harry Saini, Yam Levi, Dominik Lorenz, Axel Sauer, Frederic
  Boesel, et~al.
\newblock Scaling rectified flow transformers for high-resolution image
  synthesis.
\newblock In {\em Forty-first International Conference on Machine Learning},
  2024.

\bibitem{fan2024scaling}
Lijie Fan, Kaifeng Chen, Dilip Krishnan, Dina Katabi, Phillip Isola, and
  Yonglong Tian.
\newblock Scaling laws of synthetic images for model training... for now.
\newblock In {\em Proceedings of the IEEE/CVF Conference on Computer Vision and
  Pattern Recognition}, pages 7382--7392, 2024.

\bibitem{fergus2003object}
Robert Fergus, Pietro Perona, and Andrew Zisserman.
\newblock Object class recognition by unsupervised scale-invariant learning.
\newblock In {\em 2003 IEEE Computer Society Conference on Computer Vision and
  Pattern Recognition, 2003. Proceedings.}, volume~2, pages II--II. IEEE, 2003.

\bibitem{goodfellow2014generative}
Ian Goodfellow, Jean Pouget-Abadie, Mehdi Mirza, Bing Xu, David Warde-Farley,
  Sherjil Ozair, Aaron Courville, and Yoshua Bengio.
\newblock Generative adversarial nets.
\newblock {\em Advances in neural information processing systems}, 27, 2014.

\bibitem{he2016deep}
Kaiming He, Xiangyu Zhang, Shaoqing Ren, and Jian Sun.
\newblock Deep residual learning for image recognition.
\newblock In {\em Proceedings of the IEEE conference on computer vision and
  pattern recognition}, pages 770--778, 2016.

\bibitem{he2023synthetic}
Ruifei He, Shuyang Sun, Xin Yu, Chuhui Xue, Wenqing Zhang, Philip Torr, Song
  Bai, and Xiaojuan Qi.
\newblock Is synthetic data from generative models ready for image recognition?
\newblock In {\em ICLR}, 2023.

\bibitem{hennicke2024mind}
Leonhard Hennicke, Christian~Medeiros Adriano, Holger Giese, Jan~Mathias
  Koehler, and Lukas Schott.
\newblock Mind the gap between synthetic and real: Utilizing transfer learning
  to probe the boundaries of stable diffusion generated data.
\newblock {\em arXiv preprint arXiv:2405.03243}, 2024.

\bibitem{heusel2017gans}
Martin Heusel, Hubert Ramsauer, Thomas Unterthiner, Bernhard Nessler, and Sepp
  Hochreiter.
\newblock Gans trained by a two time-scale update rule converge to a local nash
  equilibrium.
\newblock {\em Advances in neural information processing systems}, 30, 2017.

\bibitem{ho2020ddpm}
Jonathan Ho, Ajay Jain, and Pieter Abbeel.
\newblock Denoising diffusion probabilistic models.
\newblock {\em Advances in neural information processing systems},
  33:6840--6851, 2020.

\bibitem{ho2020denoising}
Jonathan Ho, Ajay Jain, and Pieter Abbeel.
\newblock Denoising diffusion probabilistic models.
\newblock {\em Advances in Neural Information Processing Systems},
  33:6840--6851, 2020.

\bibitem{jung2024dalda}
Kyuheon Jung, Yongdeuk Seo, Seongwoo Cho, Jaeyoung Kim, Hyun-seok Min, and
  Sungchul Choi.
\newblock Dalda: Data augmentation leveraging diffusion model and llm with
  adaptive guidance scaling.
\newblock {\em arXiv preprint arXiv:2409.16949}, 2024.

\bibitem{karras2018progressive}
Tero Karras, Timo Aila, Samuli Laine, and Jaakko Lehtinen.
\newblock Progressive growing of gans for improved quality, stability, and
  variation.
\newblock In {\em ICLR}, 2018.

\bibitem{karras2022elucidating}
Tero Karras, Miika Aittala, Timo Aila, and Samuli Laine.
\newblock Elucidating the design space of diffusion-based generative models.
\newblock {\em Advances in neural information processing systems},
  35:26565--26577, 2022.

\bibitem{kingma2013auto}
Diederik~P Kingma.
\newblock Auto-encoding variational bayes.
\newblock {\em arXiv preprint arXiv:1312.6114}, 2013.

\bibitem{krizhevsky2009learning}
Alex Krizhevsky, Geoffrey Hinton, et~al.
\newblock Learning multiple layers of features from tiny images.
\newblock 2009.

\bibitem{nichol2021glide}
Alex Nichol, Prafulla Dhariwal, Aditya Ramesh, Pranav Shyam, Pamela Mishkin,
  Bob McGrew, Ilya Sutskever, and Mark Chen.
\newblock Glide: Towards photorealistic image generation and editing with
  text-guided diffusion models.
\newblock {\em arXiv preprint arXiv:2112.10741}, 2021.

\bibitem{paszke2019pytorch}
Adam Paszke, Sam Gross, Francisco Massa, Adam Lerer, James Bradbury, Gregory
  Chanan, Trevor Killeen, Zeming Lin, Natalia Gimelshein, Luca Antiga, et~al.
\newblock Pytorch: An imperative style, high-performance deep learning library.
\newblock {\em Advances in neural information processing systems}, 32, 2019.

\bibitem{Peebles2022DiT}
William Peebles and Saining Xie.
\newblock Scalable diffusion models with transformers.
\newblock {\em arXiv preprint arXiv:2212.09748}, 2022.

\bibitem{qin2023unicontrol}
Can Qin, Shu Zhang, Ning Yu, Yihao Feng, Xinyi Yang, Yingbo Zhou, Huan Wang,
  Juan~Carlos Niebles, Caiming Xiong, Silvio Savarese, et~al.
\newblock Unicontrol: A unified diffusion model for controllable visual
  generation in the wild.
\newblock {\em arXiv preprint arXiv:2305.11147}, 2023.

\bibitem{ramesh2022hierarchical}
Aditya Ramesh, Prafulla Dhariwal, Alex Nichol, Casey Chu, and Mark Chen.
\newblock Hierarchical text-conditional image generation with clip latents.
\newblock {\em arXiv preprint arXiv:2204.06125}, 1(2):3, 2022.

\bibitem{ramesh2021zero}
Aditya Ramesh, Mikhail Pavlov, Gabriel Goh, Scott Gray, Chelsea Voss, Alec
  Radford, Mark Chen, and Ilya Sutskever.
\newblock Zero-shot text-to-image generation.
\newblock In {\em International conference on machine learning}, pages
  8821--8831. Pmlr, 2021.

\bibitem{rombach2022high}
Robin Rombach, Andreas Blattmann, Dominik Lorenz, Patrick Esser, and Bj{\"o}rn
  Ommer.
\newblock High-resolution image synthesis with latent diffusion models.
\newblock In {\em Proceedings of the IEEE/CVF conference on computer vision and
  pattern recognition}, pages 10684--10695, 2022.

\bibitem{saharia2022photorealistic}
Chitwan Saharia, William Chan, Saurabh Saxena, Lala Li, Jay Whang, Emily~L
  Denton, Kamyar Ghasemipour, Raphael Gontijo~Lopes, Burcu Karagol~Ayan, Tim
  Salimans, et~al.
\newblock Photorealistic text-to-image diffusion models with deep language
  understanding.
\newblock {\em Advances in neural information processing systems},
  35:36479--36494, 2022.

\bibitem{salimans2016improved}
Tim Salimans, Ian Goodfellow, Wojciech Zaremba, Vicki Cheung, Alec Radford, and
  Xi~Chen.
\newblock Improved techniques for training gans.
\newblock {\em Advances in neural information processing systems}, 29, 2016.

\bibitem{sariyildiz2023fake}
Mert~B{\"u}lent Sar{\i}y{\i}ld{\i}z, Karteek Alahari, Diane Larlus, and Yannis
  Kalantidis.
\newblock Fake it till you make it: Learning transferable representations from
  synthetic imagenet clones.
\newblock In {\em Proceedings of the IEEE/CVF Conference on Computer Vision and
  Pattern Recognition}, pages 8011--8021, 2023.

\bibitem{schuhmann2022laion}
Christoph Schuhmann, Romain Beaumont, Richard Vencu, Cade Gordon, Ross
  Wightman, Mehdi Cherti, Theo Coombes, Aarush Katta, Clayton Mullis, Mitchell
  Wortsman, et~al.
\newblock Laion-5b: An open large-scale dataset for training next generation
  image-text models.
\newblock {\em Advances in Neural Information Processing Systems},
  35:25278--25294, 2022.

\bibitem{Seitzer2020FID}
Maximilian Seitzer.
\newblock {pytorch-fid: FID Score for PyTorch}.
\newblock \url{https://github.com/mseitzer/pytorch-fid}, August 2020.
\newblock Version 0.3.0.

\bibitem{singh2024synthetic}
Krishnakant Singh, Thanush Navaratnam, Jannik Holmer, Simone Schaub-Meyer, and
  Stefan Roth.
\newblock Is synthetic data all we need? benchmarking the robustness of models
  trained with synthetic images.
\newblock In {\em CVPR}, pages 2505--2515, 2024.

\bibitem{sohl2015deep}
Jascha Sohl-Dickstein, Eric Weiss, Niru Maheswaranathan, and Surya Ganguli.
\newblock Deep unsupervised learning using nonequilibrium thermodynamics.
\newblock In {\em International conference on machine learning}, 2015.

\bibitem{song2020denoising}
Jiaming Song, Chenlin Meng, and Stefano Ermon.
\newblock Denoising diffusion implicit models.
\newblock {\em arXiv preprint arXiv:2010.02502}, 2020.

\bibitem{song2020score}
Yang Song, Jascha Sohl-Dickstein, Diederik~P Kingma, Abhishek Kumar, Stefano
  Ermon, and Ben Poole.
\newblock Score-based generative modeling through stochastic differential
  equations.
\newblock {\em arXiv preprint arXiv:2011.13456}, 2020.

\bibitem{tian2024learning}
Yonglong Tian, Lijie Fan, Kaifeng Chen, Dina Katabi, Dilip Krishnan, and
  Phillip Isola.
\newblock Learning vision from models rivals learning vision from data.
\newblock In {\em Proceedings of the IEEE/CVF Conference on Computer Vision and
  Pattern Recognition}, 2024.

\bibitem{trabuccoeffective}
Brandon Trabucco, Kyle Doherty, Max~A Gurinas, and Ruslan Salakhutdinov.
\newblock Effective data augmentation with diffusion models.
\newblock In {\em The Twelfth International Conference on Learning
  Representations}, 2024.

\bibitem{tu2007learning}
Zhuowen Tu.
\newblock Learning generative models via discriminative approaches.
\newblock In {\em 2007 IEEE Conference on Computer Vision and Pattern
  Recognition}, 2007.

\bibitem{tu2002image}
Zhuowen Tu and Song-Chun Zhu.
\newblock Image segmentation by data-driven markov chain monte carlo.
\newblock {\em TPAMI}, 24(5):657--673, 2002.

\bibitem{van2008visualizing}
Laurens Van~der Maaten and Geoffrey Hinton.
\newblock Visualizing data using t-sne.
\newblock {\em Journal of machine learning research}, 9(11), 2008.

\bibitem{wang2024omnicontrolnet}
Yilin Wang, Haiyang Xu, Xiang Zhang, Zeyuan Chen, Zhizhou Sha, Zirui Wang, and
  Zhuowen Tu.
\newblock Omnicontrolnet: Dual-stage integration for conditional image
  generation.
\newblock In {\em Proceedings of the IEEE/CVF Conference on Computer Vision and
  Pattern Recognition}, pages 7436--7448, 2024.

\bibitem{wightman2019timm}
Ross Wightman.
\newblock Pytorch image models.
\newblock \url{https://github.com/huggingface/pytorch-image-models}, 2019.

\bibitem{yang2023medmnist}
Jiancheng Yang, Rui Shi, Donglai Wei, Zequan Liu, Lin Zhao, Bilian Ke,
  Hanspeter Pfister, and Bingbing Ni.
\newblock Medmnist v2-a large-scale lightweight benchmark for 2d and 3d
  biomedical image classification.
\newblock {\em Scientific Data}, 10(1):41, 2023.

\bibitem{yang2022image}
Suorong Yang, Weikang Xiao, Mengchen Zhang, Suhan Guo, Jian Zhao, and Furao
  Shen.
\newblock Image data augmentation for deep learning: A survey.
\newblock {\em arXiv preprint arXiv:2204.08610}, 2022.

\bibitem{yin2023ttida}
Yuwei Yin, Jean Kaddour, Xiang Zhang, Yixin Nie, Zhenguang Liu, Lingpeng Kong,
  and Qi~Liu.
\newblock Ttida: Controllable generative data augmentation via text-to-text and
  text-to-image models.
\newblock {\em arXiv preprint arXiv:2304.08821}, 2023.

\bibitem{yuille2006vision}
Alan Yuille and Daniel Kersten.
\newblock Vision as bayesian inference: analysis by synthesis?
\newblock {\em Trends in cognitive sciences}, 10(7):301--308, 2006.

\bibitem{yun2019cutmix}
Sangdoo Yun, Dongyoon Han, Seong~Joon Oh, Sanghyuk Chun, Junsuk Choe, and
  Youngjoon Yoo.
\newblock Cutmix: Regularization strategy to train strong classifiers with
  localizable features.
\newblock In {\em Proceedings of the IEEE/CVF international conference on
  computer vision}, pages 6023--6032, 2019.

\bibitem{zhang2017mixup}
Hongyi Zhang, Moustapha Cisse, Yann~N Dauphin, and David Lopez-Paz.
\newblock mixup: Beyond empirical risk minimization.
\newblock {\em arXiv preprint arXiv:1710.09412}, 2017.

\bibitem{zhang2023controlnet}
Lvmin Zhang, Anyi Rao, and Maneesh Agrawala.
\newblock Adding conditional control to text-to-image diffusion models.
\newblock In {\em Proceedings of the IEEE/CVF International Conference on
  Computer Vision}, pages 3836--3847, 2023.

\bibitem{zhou2023using}
Yongchao Zhou, Hshmat Sahak, and Jimmy Ba.
\newblock Using synthetic data for data augmentation to improve classification
  accuracy.
\newblock 2023.

\end{thebibliography}

\clearpage
\newpage

\appendix
\clearpage
\section{Details for Estimating the Equivalence Equations}\label{sec:appendix_equations}
In the main text, we introduced seven equations to estimate the quantitative equivalence between generative and real data augmentation. This section details the process of deriving the empirical equivalence. Let $n_{\text{real}}$ and $n_{\text{syn}}$ denote the total amounts of real and synthetic data used in each experiment, respectively.
Take closed-set generative data augmentation on ImageNet-100 as an example, we begin by empirically creating a set of functions:
\begin{align}
    \mathbb{H} = \{n_{\text{real}}, n_{\text{syn}}, \log{(n_{\text{real}}+\epsilon)}, \log{(n_{\text{syn}}+\epsilon)},\log{(n_{\text{real}}+n_{\text{syn}}+\epsilon)}\}.
\end{align}
where $\epsilon$ is a small constant added to avoid the logarithm approaching infinity. 
Then, we perform linear regression using all the 31 possible combinations of functions within $\mathbb{H}$ to model the relationship between Acc, $n_{\text{real}}$ and $n_{\text{syn}}$. Among these combinations, we find that the linear combination of $\log{(n_{\text{real}}+\epsilon)}, \log{(n_{\text{syn}}+\epsilon)}$ and $\log{(n_{\text{real}}+n_{\text{syn}}+\epsilon)}$ provides the best fit, i.e.
\begin{align}
\text{Acc}(n_{\text{real}}, n_{\text{syn}})=b_1\log{(n_{\text{real}}+\epsilon)}+b_2 \log{(n_{\text{syn}}+\epsilon)}+b_3\log{(n_{\text{real}}+n_{\text{syn}}+\epsilon)}+b_4.
\label{eq:acc_initial}
\end{align}  
where $b_1, b_2, b_3, b_4$ are learnable parameters.
Note that $n_{\text{real}}=\nbase+\nreal$. For each $\nbase\in\{6500, 26000\}$, we fit an equation in the form of Eq. \ref{eq:acc_initial} using the corresponding results from Tab. \ref{tab:appendix_imagenet_100_real_results} and Tab. \ref{tab:appendix_imagenet_100_C_results} (select the rows where the \textit{Base Real} entries exactly match the value of $\nbase$). 
    After deriving this functional relationship, we can visualize the equivalence curves, as shown in Fig. 2(c) in the main text. Then, based on Eq. \ref{eq:acc_initial}, we numerically solve for a set of tuples: \begin{align}
    \mathbb{E}=\{
    (\nbase, \nreal, \nsyn) \mid \text{Acc}(\nbase+\nreal, 0)=\text{Acc}(\nbase, \nsyn)\}.
    \end{align}
    
    The values chosen for $\nbase$ and $\nreal$ in $\mathbb{E}$ are as follows:

    \begin{itemize}[itemsep=1.5pt]
    \item CIFAR-10: $\nbase \in \{500, 5000, 25000\}$. At each $\nbase$, $\frac{\nreal}{\nbase} \in \{1.0, 2.0, 3.0\}$.

    \item ImageNet-100: $\nbase \in \{6500, 26000\}$. At each $\nbase$, $\frac{\nreal}{\nbase} \in \{0.2, 0.4, 0.6, 0.8, 1.0, 2.0,$ $ 3.0, 4.0\}$.

    \item BloodMNIST: $\nbase \in \{1200, 6000\}$. At each $\nbase$, $\frac{\nreal}{\nbase} \in \{1.0, 2.0, 3.0\}$.
    
    \end{itemize}

    For each $\nbase$ and $\nreal$, we solve for $\nsyn$ using the equation $\text{Acc}(\nbase+\nreal, 0) = \text{Acc}(\nbase, \nsyn)$ and obtain the tuple $(\nbase, \nreal, \nsyn)$. If $\frac{\nsyn}{\nbase}$ exceeds 100 for a given tuple, we discard it to ensure stable equivalence fitting in the next step.
    
    Subsequently, we assume the quantitative equivalence is in the form of: \begin{equation}
    {\frac{\nsyn}{\nbase} \simeq c_1^{\tau\nbase}\times \left(c_2^{\frac{\nreal}{\nbase}}-1 \right)}.
    \label{eq:esti_initial}
    \end{equation}
    and regress the parameters $c_1, c_2$ from the tuples in $\mathbb{E}$. We follow the same procedures to derive the equations on CIFAR-10 and BloodMNIST. Since $n_{\text{base}}$ is typically in the thousands, we introduce a constant $\tau$ to balance the calculation and prevent $c_2$ from approaching 1 asymptotically. In practice, given that the total size of the original training set is $n_{\text{total}}$, we set $\tau =\frac{10}{n_\text{total}}$ (e.g., for CIFAR-10, which has 50,000 images, $\tau =\frac{1}{5000}$).
    
    In the main text, we used the parameterized forms of this equation to validate a series of conclusions. Although it may appear ad-hoc, its consistency with patterns observed in the experimental results demonstrates its validity. For this part, the equation numbers refer to those in the main text.
    \begin{itemize} [itemsep=5pt]
    \item The value of $c_1$ in \textbf{Eq. \ref{eq:esti_external_cifar}} is larger than 1, so if we \textbf{fix} $\frac{\nreal}{\nbase}=\bf2$, the equivalent $\frac{\nsyn}{\nbase}$ should \textbf{go up} as $\nbase$ increases. As observed in Tab. \ref{tab:appendix_cifar_results_real} and Tab. \ref{tab:appendix_cifar_cifake_results}, at $\nbase=500$, the equivalent $\frac{\nsyn}{\nbase}\in (3,4)$. At $\nbase=5,000$, the equivalent $\frac{\nsyn}{\nbase}$ is larger than 5, which means that for open-set generative data augmentation, achieving further accuracy gains becomes more challenging when the real data amount is already large.
    \item The value of $c_4$ in \textbf{Eq. \ref{eq:esti_external_imagenet}} is larger than 1, meaning that if we \textbf{fix} $\nbase=\bf6,500$, $\frac{\nsyn}{\nreal}$ should \textbf{go up} as $\nreal$ increases. As shown in Tab. \ref{tab:appendix_imagenet_100_real_results} and Tab. \ref{tab:appendix_imagenet_100_C_results}, at $\nreal=0.5\times\nbase$, the equivalent $\nsyn<1\times\nbase$, leading to $\frac{\nsyn}{\nreal}<2$. At $\nreal=4\times\nbase$, the equivalent $\nsyn>10\times\nbase$, resulting in $\frac{\nsyn}{\nreal}>2.5$, which means synthetic data has a diminishing marginal effect.
    \end{itemize}

\section{Implementation Details}
\label{Appendix: Impl.}

\begin{table}
\caption{Iterations of the ImageNet-10 and ImageNet-100 experiments with respect to data amount.}\label{tab: appendix_iterations}
  \vskip 0.05in
  \centering
  \begin{threeparttable}
  \begin{small}
  \renewcommand{\multirowsetup}{\centering}
  \setlength{\tabcolsep}{2pt}
  \renewcommand{\arraystretch}{1.2}
  \begin{tabular}{c|cccc}
    \toprule
    {Data amount} & 
   [650, 1.3k) & [1.3k, 2.6k) & [2.6k, 13k) & [13k, 260k] \\
    \midrule
    {Iterations}
    & 10k & 30k & 60k & 120k \\
    \bottomrule
  \end{tabular}
    \end{small}
  \end{threeparttable}
  
\end{table}

\begin{table}
  \caption{Details of the hyperparameters.}
  \label{tab: appendix_supervised_hyper}
  \vskip 0.05in
  \centering
  \begin{threeparttable}
  \begin{small}
  \renewcommand{\multirowsetup}{\centering}
  \setlength{\tabcolsep}{4pt}
  \renewcommand{\arraystretch}{1.2}
  \begin{tabular}{l|ccc}
    \toprule
    {Hyper-parameter} & 
   {ResNet-50} & ViT-B/32 & ResNet-110 \\
    \midrule
    {Batch size}
    & 192 & 192 & 128 \\
    \multirow{2}{*}{Base lr} & \multirow{2}{*}{0.1} & \multirow{2}{*}{0.1} & CIFAR-10: 0.1 \\
    & & & BloodMNIST: 0.01 \\
    Decay method & cosine & cosine & multistep \\
    Optimizer & SGD & SGD & SGD\\
    Momentum & 0.9 & 0.9 & 0.9\\
    Weight decay & 1e-4 & 0 & 1e-4 \\
    Warmup iterations & 10\% & 10\% & no warmup \\

    \bottomrule
  \end{tabular}
    \end{small}
  \end{threeparttable}
\end{table}

For ImageNet-10 and ImageNet-100 experiments, we fix the number of total iterations instead of total epochs. However, since our mixed dataset size varies significantly, ranging from 650 to 260,000, we group the total number of iterations into different levels based on the data size, and the details are provided in Tab. \ref{tab: appendix_iterations}. All images are resized to 224 $\times$ 224 for input. For data augmentation, we apply random cropping, resizing, and random horizontal flipping. The above settings hold for both ResNet-50 and ViT-B/32 experiments.

For the CIFAR-10 and BloodMNIST experiments, we use ResNet-110 as introduced in \citep{he2016deep} for classification, as it is well-suited for the smaller image sizes of CIFAR-10. In the experiments on CIFAR-10-Mid, CIFAR-10-Large, BloodMNIST-Mid and BloodMNIST-Large, we fix the number of epochs at 160, using a multistep scheduler to decay the learning rate by a factor of 0.1 at epochs 80 and 120. For CIFAR-10-Small and BloodMNIST-Small, the training lasts for 320 epochs, and the learning rate decays at epoch 160 and 240. For CIFAR-10-Tiny, the training lasts for 480 epochs, and the learning rate decays at epoch 240 and 360. For data augmentation, we apply random cropping and random horizontal flipping. 

Notably, BloodMNIST is an imbalanced dataset. When selecting its subset or performing real data augmentation, we ensure that the class proportions remain consistent. However, for generative data augmentation, we leverage the ability of generative models to produce any amount of synthetic data, adding an equal number of synthetic images to each class to make the dataset more balanced.

For all settings mentioned above, we run \textbf{3} trials for each experiment and report the average result. Details on other hyper-parameters are provided in Tab. \ref{tab: appendix_supervised_hyper}. 

In our mixed training approach, we utilize PyTorch's \citep{paszke2019pytorch} \texttt{ConcatDataset} method to combine real and synthetic data. The \texttt{RandomSampler} of PyTorch randomly shuffles the combined dataset at the start of each epoch.

When calculating the Frechet Inception Distance (FID) score between synthetic and real datasets, we employ the official PyTorch implementation of FID \citep{Seitzer2020FID}.

\begin{figure*}[t]
\begin{center}
\centerline{\includegraphics[width=\textwidth]{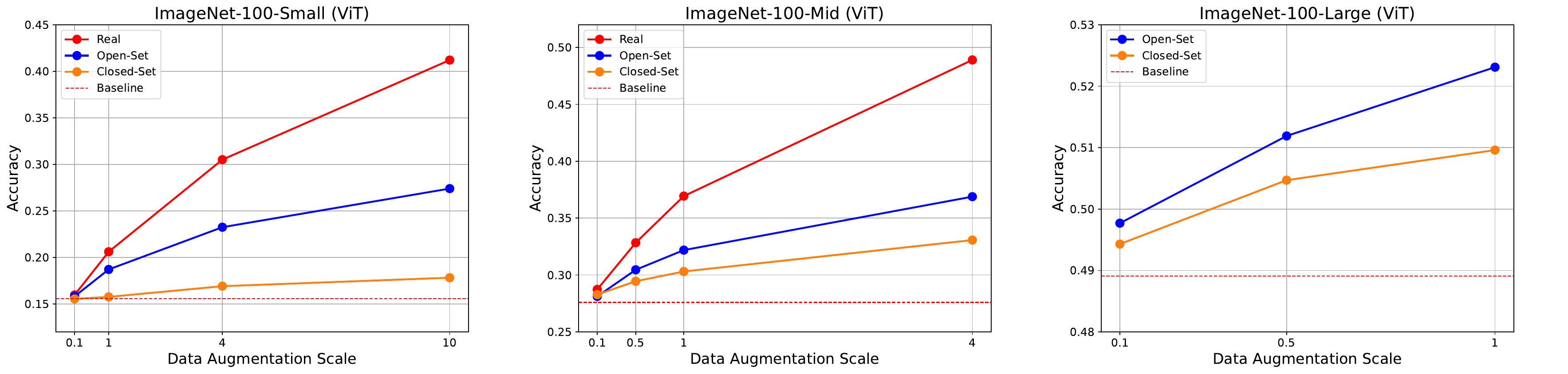}}
	\caption{Accuracy curves on ImageNet-100 with ViT-B/32 as the classifier.}
	\label{fig:appendix_vit}
\end{center}
\vspace{-20pt}
\end{figure*} 

\section{Classification Results for Generalization Studies on ViT-B/32}\label{sec:appendix_ablation}

To demonstrate the generalizability of our conclusions, we replaced ResNet-50 \citep{he2016deep} with ViT-B/32 \citep{dosovitskiy2020image} as the classifier on ImageNet-100. Fig. \ref{fig:appendix_vit} presents the classification results under varying base training set sizes and generative data augmentation sizes. 

\begin{table}[h]
\caption{Accuracy comparison of traditional and generative augmentation settings.}
  \label{tab:traditional_aug}
  \vskip 0.05in
  \centering
  \begin{threeparttable}
  \begin{small}
  \renewcommand{\multirowsetup}{\centering}
  \setlength{\tabcolsep}{7pt}
  \begin{tabular}{c|c|c|c}
    \toprule
   \multirow{2}{*}{{Real Dataset}} &{Traditional} & {Generative} &
    \multirow{2}{*}{Accuracy} \\
     
     & Augmentation & {Augmentation} &\\
    \toprule
    
     \multirow{4}{*}{{CIFAR-10}} &{{\ding{55}}} & \ding{55} & {{85.00}} \\
     & RandomHorizontalFlip & \ding{55} & 89.04 \textcolor{mygreen}{\text{\scriptsize (+4.04)}} \\
     & RandomCrop & \ding{55} & {89.90} \textcolor{mygreen}{\text{\scriptsize (+4.90)}} \\
     & \ding{55} & EDM \citep{karras2022elucidating} & {\bf 90.53} \textcolor{mygreen}{\text{\scriptsize (+5.53)}} \\
    \bottomrule
  \end{tabular}
    \end{small}
  \end{threeparttable}
    \vspace{-6pt}
\end{table}

\section{Comparison with Traditional Data Augmentation}\label{appendix:traditional augmentation}
We conduct additional experiments on CIFAR-10-Large to compare the effectiveness of generative and traditional data augmentation. Specifically, we first conduct a baseline experiment without any data augmentation. In addition, we perform three separate experiments, each applying a single augmentation method: Random Cropping, Random Horizontal Flipping, or closed-set generative data augmentation. For closed-set generative data augmentation, we enhance the base training set by incorporating synthetic data generated by an EDM model \citep{karras2022elucidating} trained on CIFAR-10-Large, maintaining a real-to-syn data ratio of 1:1. The experiment results are shown in Tab. \ref{tab:traditional_aug}. We find that on CIFAR-10-Large, generative data augmentation is more effective than applying a single traditional data augmentation method, highlighting the research value of this approach.

It is worth noting that, apart from this set of experiments, all other experiments on CIFAR-10 utilize the full set of traditional data augmentation methods, including Random Cropping and Random Horizontal Flipping.

\section{ImageNet-10 Dataset Description}\label{appendix:imagenet-10}
\begin{figure*}[t]
\begin{center}
\centerline{\includegraphics[width=\textwidth]{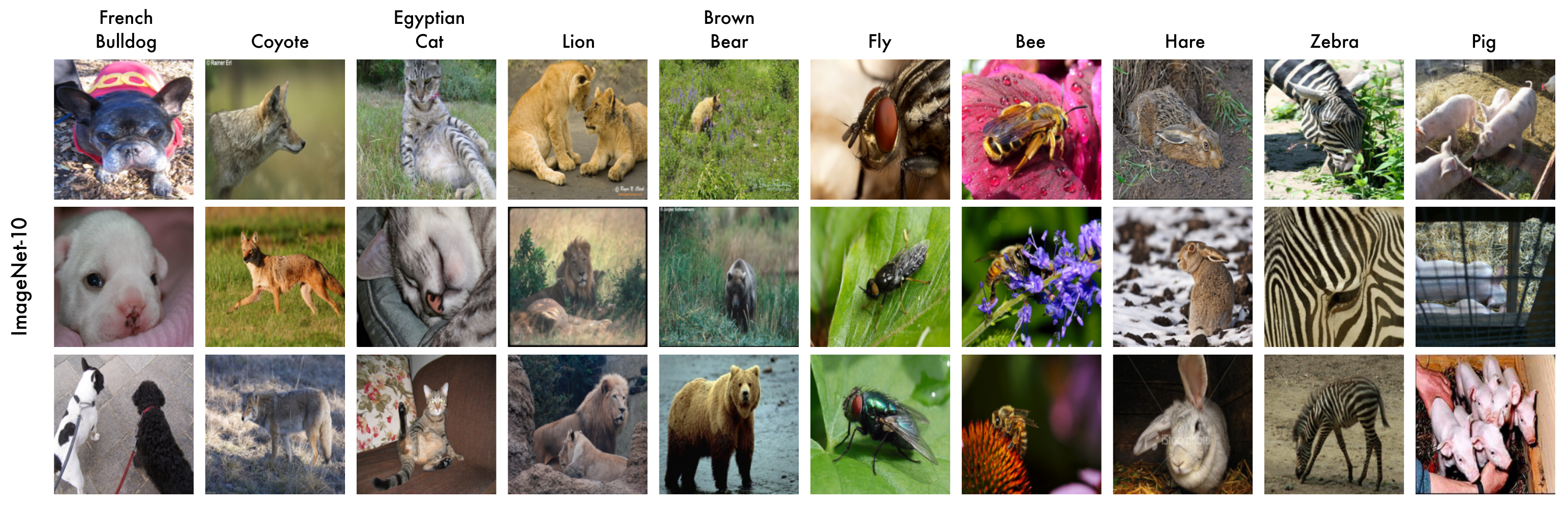}}
    \vspace{-5pt}
	\caption{Visualizations of ImageNet-10. Ten classes are randomly selected from the original ImageNet-1k.}
	\label{fig:appendix_vis_imagenet-10}
\end{center}
\vspace{-20pt}
\end{figure*}

In the following experiments, we will use the ImageNet-10 dataset. Before proceeding, we provide a brief introduction to it: ImageNet-10 comprises 10 classes of animals selected from the original ImageNet-1k \citep{5206848}. Each class contains 1,300 images. The class labels are French bulldog, coyote, Egyptian cat, lion, brown bear, fly, bee, hare, zebra, and pig. The visualization of ImageNet-10 is provided in Fig. \ref{fig:appendix_vis_imagenet-10}.

\section{The Fundamental Difference of Real and Synthetic Data}\label{appendix:fundamentally_different}

In this section, we provide more detailed information about the experiments mentioned in Section 3.

To investigate the differences in data distribution, we conduct extensive experiments across various datasets using different synthetic data generation methods. A domain classifier is trained to distinguish between input images from the real domain and the synthetic domain. Considering both global and local statistical differences, we evaluate two scenarios: one where the input images are original images and another where the input images are image patches. 

\begin{figure*}[t]
\centering
\includegraphics[width=\linewidth]{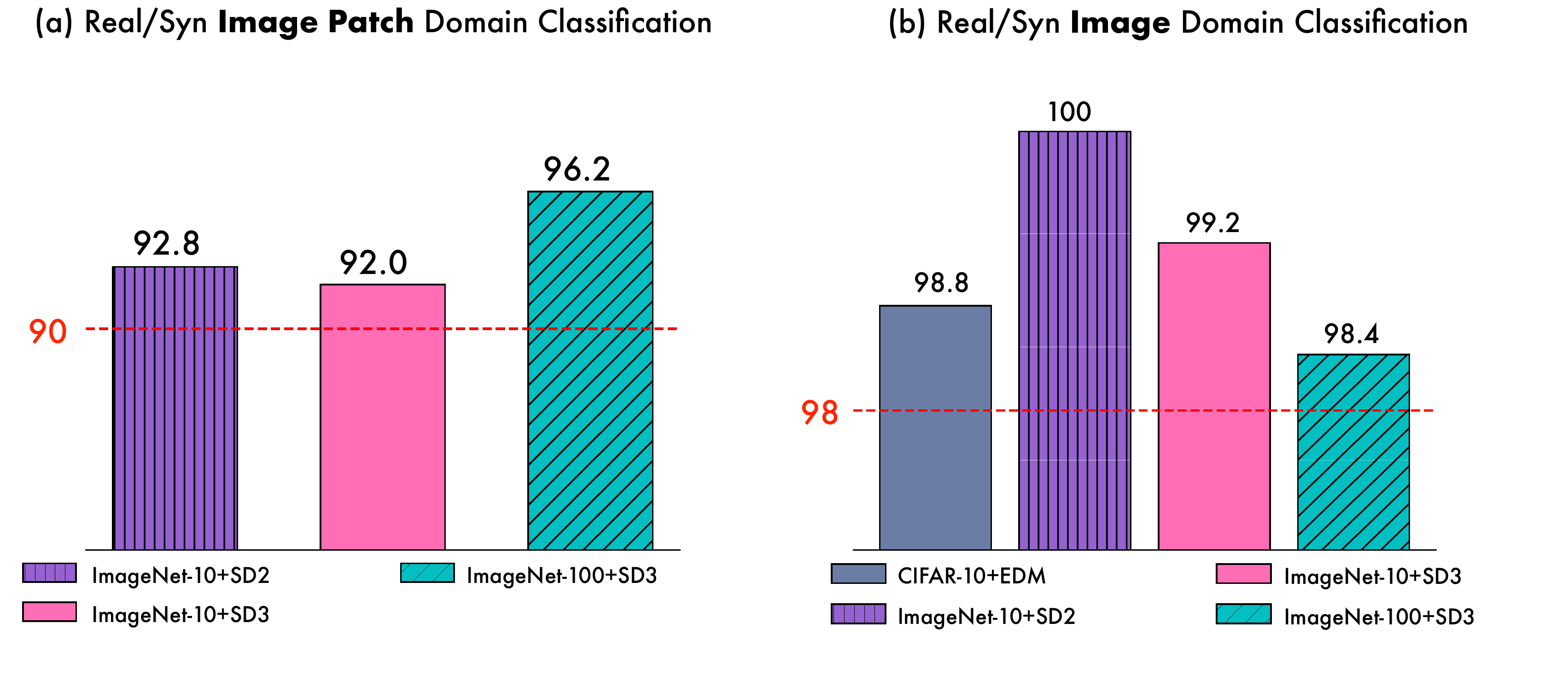}
    \vspace{-10pt}
	\caption{Visualizations of the domain classification experiments on image patches (a) and original images (b). The numbers in the figure are the accuracy ($\%$).}
	\label{fig:appendix_divide}
\end{figure*}

The real datasets include CIFAR-10, ImageNet-10, and ImageNet-100 training set. For CIFAR-10, we utilize datasets generated by EDM \citep{karras2022elucidating}, as introduced in previous sections, as synthetic datasets. Given the image resolution of these datasets is \(32 \times 32\), we only consider scenarios where the input images are the original images. For ImageNet-10, we employ ImageNet-10-SD2 and ImageNet-10-SD3 as synthetic datasets, conducting experiments on both original images and image patches. For ImageNet-100, we use ImageNet-100-SD3. All synthetic datasets are the same size as the corresponding real dataset. The results are presented in Fig. \ref{fig:appendix_divide}. All experiments involving the classification of image patches achieve an accuracy greater than $90\%$, while those involving the classification of full images achieve an accuracy exceeding $98\%$. These findings demonstrate a fundamental difference in the distribution between real and synthetic data.


\section{Addition Experiments on Open-Set Generative Data Augmentation}
In this section, we discuss several key topics related to the commonly used open-set generative data augmentation. First, we examine the impact of synthetic data quality on classification performance. Second, we explore the use of open-set generative augmentation to assist zero-shot image classification.

\begin{table}[t]
\caption{Detailed configurations of the synthetic datasets used for image classification on ImageNet-10. Data quality is measured by FID (vs. ImageNet-10 training set) and Inception Score (IS).}
\label{tab:quality_comparison}
  \centering
  \vskip 0.05in
  \begin{threeparttable}
  \begin{small}
  \renewcommand{\multirowsetup}{\centering}
  \setlength{\tabcolsep}{5pt}
  \begin{tabular}{c|c|c|c|c|cc}
    \toprule
    {Generative Model} & 
   {\rotatebox{0}{\scalebox{1.0}{ Data Amount}}} &
   {CFG Scale}&
   {Prompt} &
   {\# Classes }& 
    {FID$\downarrow$}  & {IS$\uparrow$} \\
    \toprule
    {Stable Diffusion 2}
    & 130k & 7.5 & ``High-quality photo of a $c$.'' & 10 & 40.54 & \scalebox{1.0}{1.37}\\
    Stable Diffusion 3 &130k & 2.0 & Generated captions \citep{tian2024learning} & 10  & {{\bf24.79}} & {{\bf8.30}}  \\
    
    \bottomrule
  \end{tabular}
    \end{small}
  \end{threeparttable}
    
\end{table}

\subsection{The Impact of Synthetic Data Quality on Classification Performance}

\paragraph{Experiment Setup} 
The following experiments are conducted on ImageNet-10 using open-set generative data augmentation. To generate synthetic images of varying quality, we employ two different generation protocols. The first protocol uses Stable Diffusion 2 \citep{ramesh2022hierarchical} with a straightforward class-conditioned prompt of the form $p_c=$ ``High-quality photo of a $c$'', where $c$ represents the class name. The second protocol uses Stable Diffusion 3 \citep{esser2024scaling} with diverse captions \citep{tian2024learning}, as described in the main text. We denote the two generated datasets as \textbf{ImageNet-10-SD2} and \textbf{ImageNet-10-SD3}, respectively. We then fix the synthetic data size at 1,300 images per class and add real data at a syn-to-real ratio ranging from 1:0.01 to 1:1.

\begin{wrapfigure}{r}{0.4\textwidth}
\begin{center}
  \vspace{-30pt}
\includegraphics[width=0.33\textwidth]{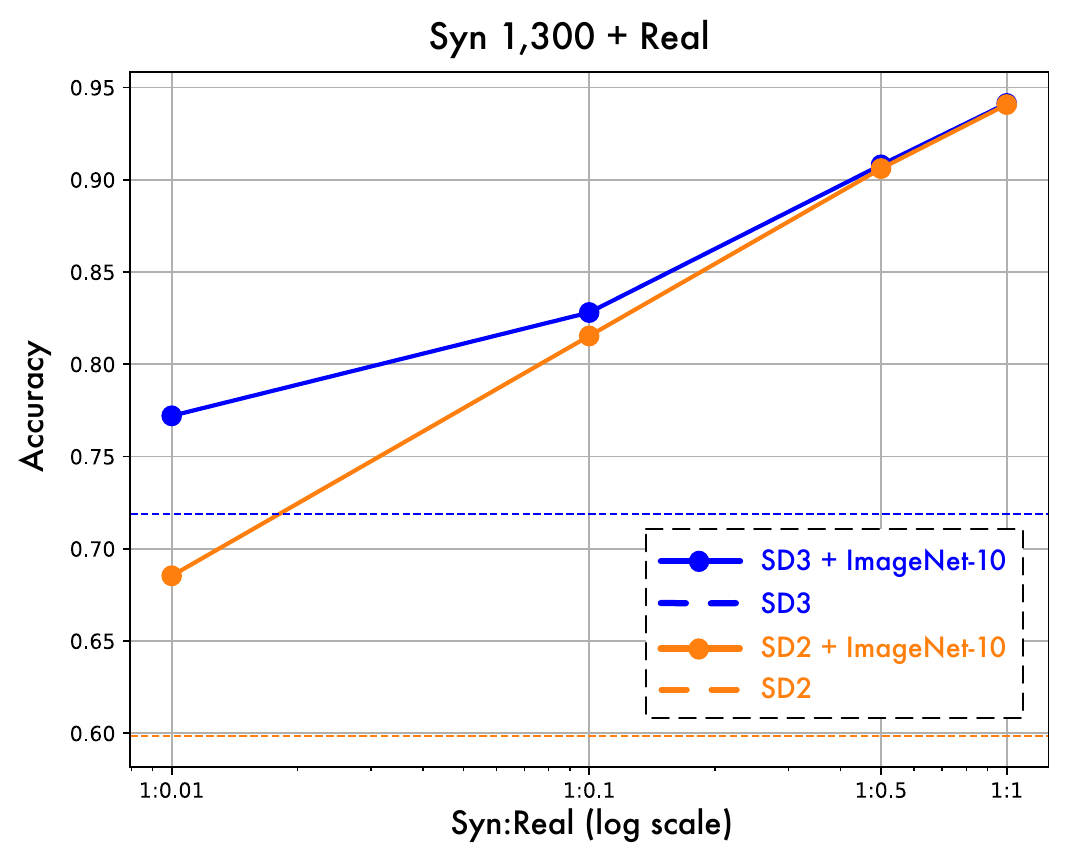}
\end{center}
 \vspace{-10pt}
	\caption{{\bf Accuracy comparison} by adding real data to synthetic data of different qualities.}
 \vspace{-20pt}
 \label{fig:appendix_quality}
\end{wrapfigure}

 \vspace{10pt}
\paragraph{Main Results}

From Tab. \ref{tab:quality_comparison}, we observe that using rich captions as prompts for Stable Diffusion 3 generation leads to significantly higher quality synthesized data compared to using Stable Diffusion 2 and simple prompts. The classification results are illustrated in Fig. \ref{fig:appendix_quality}. When only synthetic data is used, the model trained on the SD3-generated dataset achieves an accuracy of 71.87\%, which is \textbf{12.00\%} higher. However, as more real data is added into the training set, the accuracy gap between the synthetic datasets gradually closes. When the amount of real data matches the synthetic data, the difference narrows to just \textbf{0.06\%}. Therefore, we conclude that {\bf the quality of synthetic data matters more when there is less amount of real data.}

\subsection{Open-Set Generative Data Augmentation for Zero-shot Image Classification}

\paragraph{Preliminaries}

In the experiments below, we explore open-set generative data augmentation in the context of zero-shot image classification, a scenario where closed-set methods fall short. First, we try to formally define this problem.

Zero-shot image classification can be formalized as follows: Let $\mathcal{Y}_\text{train}$ be the set of training categories, the training set $\mathcal{D}_\text{train}$ consists of samples $\{(x_i,y_i)\}_{i=1}^N$, with $x_i$ representing an image and $y_i\in \mathcal{Y}_\text{train}$ its class label. The validation set $\mathcal{D}_\text{test}$ includes samples $\{(x_j, y_j)\}_{j=1}^M$, where $x_j$ is a test image, but its class label $y_j \in \mathcal{Y}_\text{test}$ is not part of the training data, i.e., $\mathcal{Y}_\text{train} \cap \mathcal{Y}_\text{test} = \emptyset$. During training, the class label $y$ is first mapped to a text description $M(y)$, where $M\in \mathcal{M}$ is a natural language template. Then, a text encoder $T$ converts $M(y)$ into a feature vector, which is subsequently projected onto the joint embedding space using a linear layer, resulting in a text embedding $\mathrm{Emb}_\text{text}(y)$. For images, a visual encoder processes the image $x$ to produce its feature representation $I(x)$. This feature is also projected onto the same space as the text embedding. The training goal is to maximize the cosine similarity between $\mathrm{Emb}_\text{text}(y)$ and $\mathrm{Emb}_\text{image}(x)$. In the testing period, given an image $x_\text{test}$, it is classified into the category $\hat y_\text{test}$ with the highest similarity score. Although the categories of the test images are not seen during training, this approach enables classification by leveraging their semantic relationships with seen classes.

In our case, by leveraging the capabilities of a open-set generative model, we can create a synthetic dataset $\mathcal{D}_\text{syn}$ composed solely of images of $\mathcal{Y}_\text{test}$ and mix it with the original training set $\mathcal{D}_\text{real}$, resulting in a combined training set $\mathcal{D}_\text{mixed}=\mathcal{D}_\text{real}\cup \mathcal{D}_\text{syn}$. In this way, $\mathcal{Y}_\text{test}$ is included in the categories of $\mathcal{D}_\text{mixed}$, while the classifier has only not seen the \textbf{real} data of $\mathcal{Y}_\text{test}$. 

\paragraph{Experiment Setup} 
In the zero-shot setting, we split ImageNet-10 and ImageNet-100 into two subsets, each containing 5 and 50 categories, respectively. We will use ImageNet-100 as an example to introduce the experiment setting, with ImageNet-10 following a similar approach. The first subset of the training set is used as the real training data, while the second subset of the validation set is reserved for testing. This ensures that the model is never exposed to the real data from the categories in the validation set. To incorporate synthetic data, we apply the same split on ImageNet-100-SD3 and only retain the second subset. This is equivalent to leveraging the generative model to produce data for the test categories. With SD3-generated synthetic images, we can align all 100 classes during training. We use a pre-trained and frozen BERT \citep{devlin-etal-2019-bert} as the text encoder and train a ResNet-50 model \cite{he2016deep} as the image encoder from scratch. Text and image features are projected onto a joint embedding space with a dimension of 512. The training goal is to maximize the cosine similarity between the text and image embeddings of the same categories. In the evaluation period, a test image is classified into the category with the highest similarity score. We keep the number of real images fixed at 1,300 per class and vary the proportion of synthetic data.
\begin{wrapfigure}{r}{0.33\textwidth}
\begin{center}
    \vspace{-10pt}\includegraphics[width=0.33\textwidth]{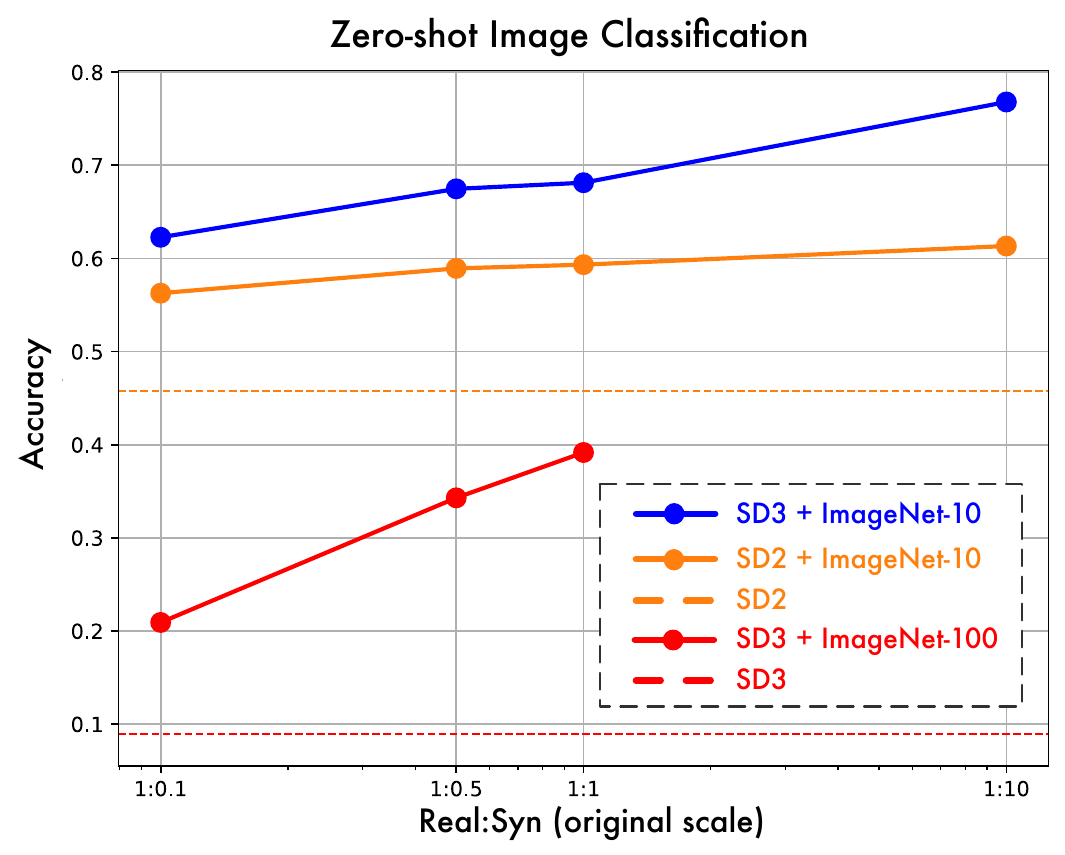}
\end{center}
 \vspace{-5pt}
	\caption{{\bf Zero-shot classification accuracy} w.r.t. the synthetic data ratio on ImageNet-10 and ImageNet-100. The size of real data is fixed at 1,300 per class.}
    \vspace{-40pt}
	\label{fig:open-world}
\end{wrapfigure}

\paragraph{Main Results}\vspace{-7pt} The results are shown in Fig. \ref{fig:open-world}. 
We find that adding SD3-generated images for the test categories, with just 0.1 times the amount of real data, improves accuracy by \textbf{17.54\%} on ImageNet-10 and \textbf{11.99\%} on ImageNet-100. Moreover, the further improvement is still notable when the synthetic data ratio reaches 1:1, demonstrating the significant potential of synthetic data in zero-shot classification. We also observe that the quality of synthetic data plays a more crucial role in the zero-shot setting, as SD3-generated images achieve \textbf{15.47\%} higher accuracy at the ratio of 1:10. Our findings align with what has been observed when the amount of real data is minimal in the supervised setting. We suggest that, since zero-shot classification lacks any real data from the test categories, it can be considered a natural extension of supervised learning with diminishing real data, which explains why these results are logical and expected.

\section{Computational Costs of Synthetic Data Generation}\label{sec:appendix_computational_cost}
In this section, we present the computational resources used in our experiments, including memory usage of sampling and GPU hours, as detailed in Tab. \ref{tab:computational_cost}. Notably, generating the same scale of synthetic data for ImageNet-100 using Stable Diffusion 3 requires 4 A6000 GPUs running for approximately 3 days.

\begin{table*}
\vspace{-10pt}
\caption{Computational costs of generating synthetic datasets used in our experiments.}
  \label{tab:computational_cost}
  \centering
  \vskip 0.05in
  \begin{threeparttable}
  \begin{small}
  \renewcommand{\multirowsetup}{\centering}
  \setlength{\tabcolsep}{6pt}
  \begin{tabular}{c|c|c|c|c|c|c}
    \toprule
    \multirow{2}{*}{Real Dataset} &  \multirow{2}{*}{Machine}&
    \multirow{2}{*}{Generative Model}& Sampling &
   Batch& 
    \multirow{2}{*}{Memory Usage}  & Time \\
     & & & Steps & Size & & per image \\
    \toprule
    CIFAR-10 / & \multirow{2}{*}{A4500}
    & \multirow{2}{*}{EDM} & \multirow{2}{*}{18} & \multirow{2}{*}{100} & \multirow{2}{*}{4,000 MB}  & \multirow{2}{*}{0.06s}\\
    BloodMNIST &
    & & & & & \\
    \cmidrule(lr){1-7}
    \multirow{3}{*}{ImageNet-10/100} & \multirow{3}{*}{A6000} & Stable Diffusion 2 & 50 & 1 & 3,900 MB  
& 1.0s \\
& & Stable Diffusion 3 & 28 & 1 & 21,800 MB  & 7.0s\\
& & DiT & 250 & 1 & 9,000 MB  & 3.5s\\
    
    \bottomrule
  \end{tabular}
    \end{small}
  \end{threeparttable}
  \vspace{-8pt}
\end{table*}

\section{Assets Sources}
\label{appendix: assets}
We summarize all external assets used in this work, including datasets, codebases, and pretrained models.

\begin{itemize}
  \item \textbf{Codebases}
  \begin{itemize}
    \item \texttt{\href{https://github.com/hysts/pytorch_image_classification}{pytorch\_image\_classification}} (MIT License)
    \item \texttt{\href{https://github.com/jeonsworld/ViT-pytorch}{vit\_pytorch}} (MIT License)
    \item \texttt{\href{https://github.com/google-research/syn-rep-learn}{syn-rep-learn}} \cite{tian2024learning} (Apache 2.0 License)
    \item \texttt{\href{https://github.com/NVlabs/edm?tab=readme-ov-file}{EDM}} \citep{karras2022elucidating} (NVIDIA Source Code License)
    \item \texttt{\href{https://github.com/facebookresearch/DiT}{DiT}} \citep{Peebles2022DiT} (CC BY-NC 4.0 License)
    \item \texttt{\href{https://github.com/CVMI-Lab/SyntheticData}{SyntheticData}} \cite{he2023synthetic} (Apache 2.0 License)
  \end{itemize}

  \item \textbf{Models}
  \begin{itemize}
    \item \texttt{\href{https://huggingface.co/stabilityai/stable-diffusion-2}{Stable Diffusion 2}} \cite{rombach2022high} (CreativeML Open RAIL++-M License)
    \item \texttt{\href{https://huggingface.co/stabilityai/stable-diffusion-3-medium}{Stable Diffusion 3}} \cite{esser2024scaling}
  \end{itemize}

  \item \textbf{Datasets}
  \begin{itemize}
    \item \texttt{\href{https://www.cs.toronto.edu/~kriz/cifar.html}{CIFAR-10}} \cite{krizhevsky2009learning}
    \item \texttt{\href{https://www.image-net.org/challenges/LSVRC/2012/index.php}{ImageNet-1k}} \cite{deng2009imagenet}
    \item \texttt{\href{https://www.kaggle.com/datasets/birdy654/cifake-real-and-ai-generated-synthetic-images}{CIFAKE}} \cite{bird2024cifake} (MIT License)
    \item \texttt{\href{https://zenodo.org/records/10519652}{BloodMNIST}} \cite{yang2023medmnist} (CC BY 4.0)
  \end{itemize}
\end{itemize}

\section{Addition Visualizations}\label{appendix_vis}
\subsection{Synthetic Data for BloodMNIST}
\begin{figure*}[!h]
\begin{center}
\centerline{\includegraphics[width=\textwidth]{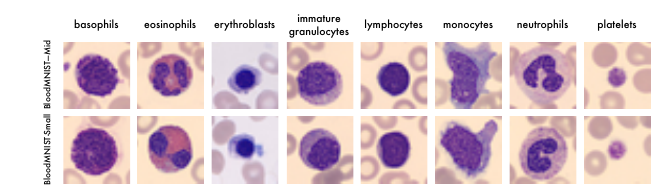}}
    \vspace{-5pt}
	\caption{Visualizations of the EDM-generated synthetic dataset for BloodMNIST-Mid and BloodMNIST-Small.}
	\label{fig:appendix_medical_mid_small}
\end{center}

\end{figure*}

We visualize the synthetic datasets for BloodMNIST-Mid and BloodMNIST-Small generated by EDM. The visualizations are shown in Fig. \ref{fig:appendix_medical_mid_small}.
\subsection{Synthetic Data for CIFAR-10}\label{sec:appendix_cifar-10_syn_data}
\begin{figure*}[h]
\begin{center}
\centerline{\includegraphics[width=0.9\textwidth]{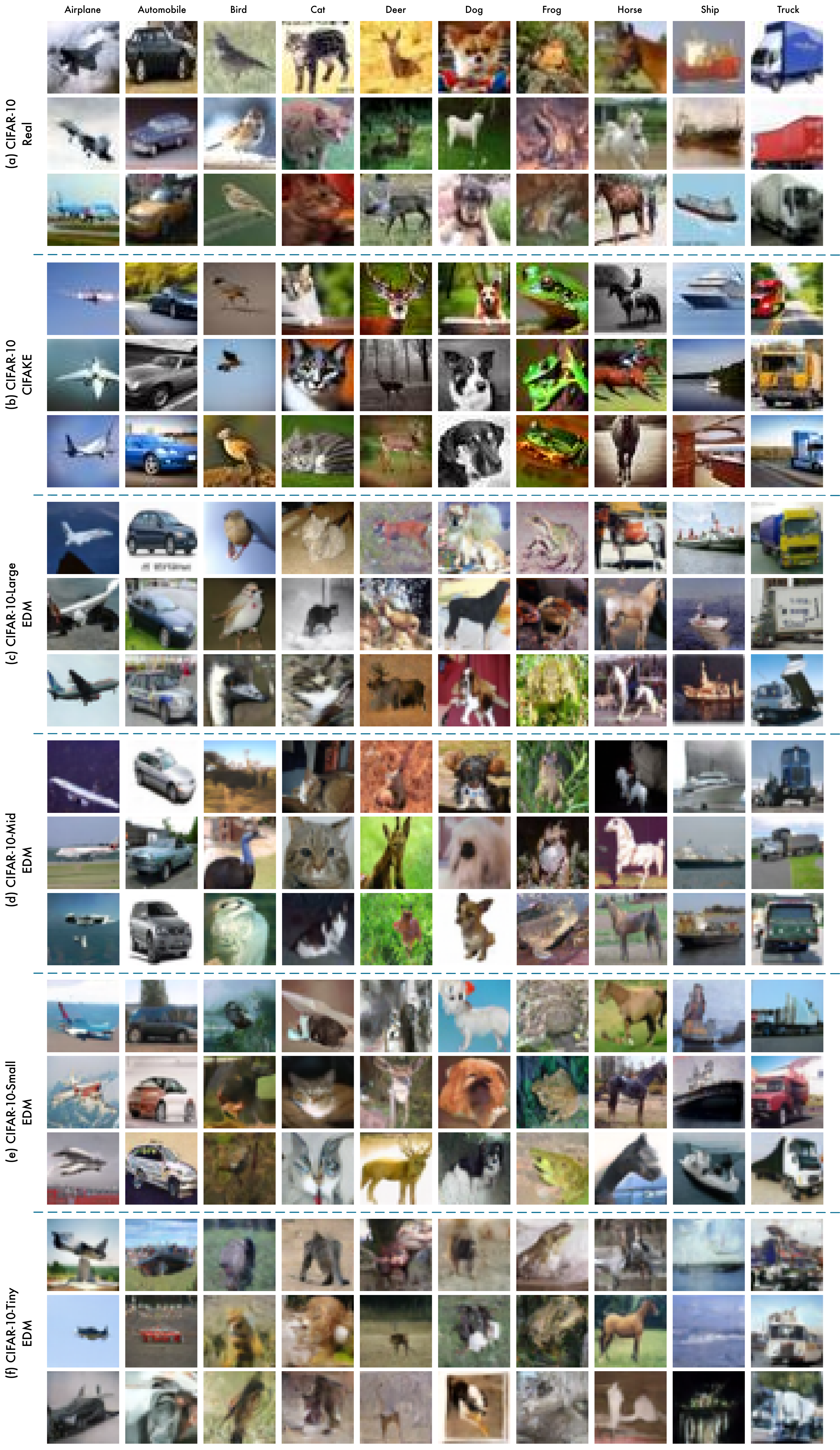}}
	\caption{Visualizations of CIFAR-10 and the closed-set synthetic datasets for CIFAR-10-Large, CIFAR-10-Mid, CIFAR-10-Small, and CIFAR-10-Tiny, as well as the open-set synthetic dataset CIFAKE.}
	\label{fig:appendix_vis_cifar}
\end{center}

\end{figure*}

We visualize some example images of the synthetic datasets for CIFAR-10, including the datasets generated by EDM, which is trained on different subsets of CIFAR-10, and the synthetic dataset from CIFAKE \citep{bird2024cifake}. The visualizations are shown in Fig. \ref{fig:appendix_vis_cifar}. 

Qualitatively, both the EDM-generated dataset (trained on full scale CIFAR-10) and CIFAKE have relatively high recognizability. However, CIFAKE images exhibit domain shifts: For example, the ship in the third row in {Fig. \ref{fig:appendix_vis_cifar} (b)} is generated as an interior scene rather than its external form. The image quality declines for EDM models trained on smaller datasets. When trained on CIFAR-10-Tiny, certain images, such as those of cats and dogs, become almost unrecognizable to the human eye.

\subsection{Synthetic Data for ImageNet-10 and ImageNet-100}

\begin{figure*}[h]
\begin{center}
\centerline{\includegraphics[width=\textwidth]{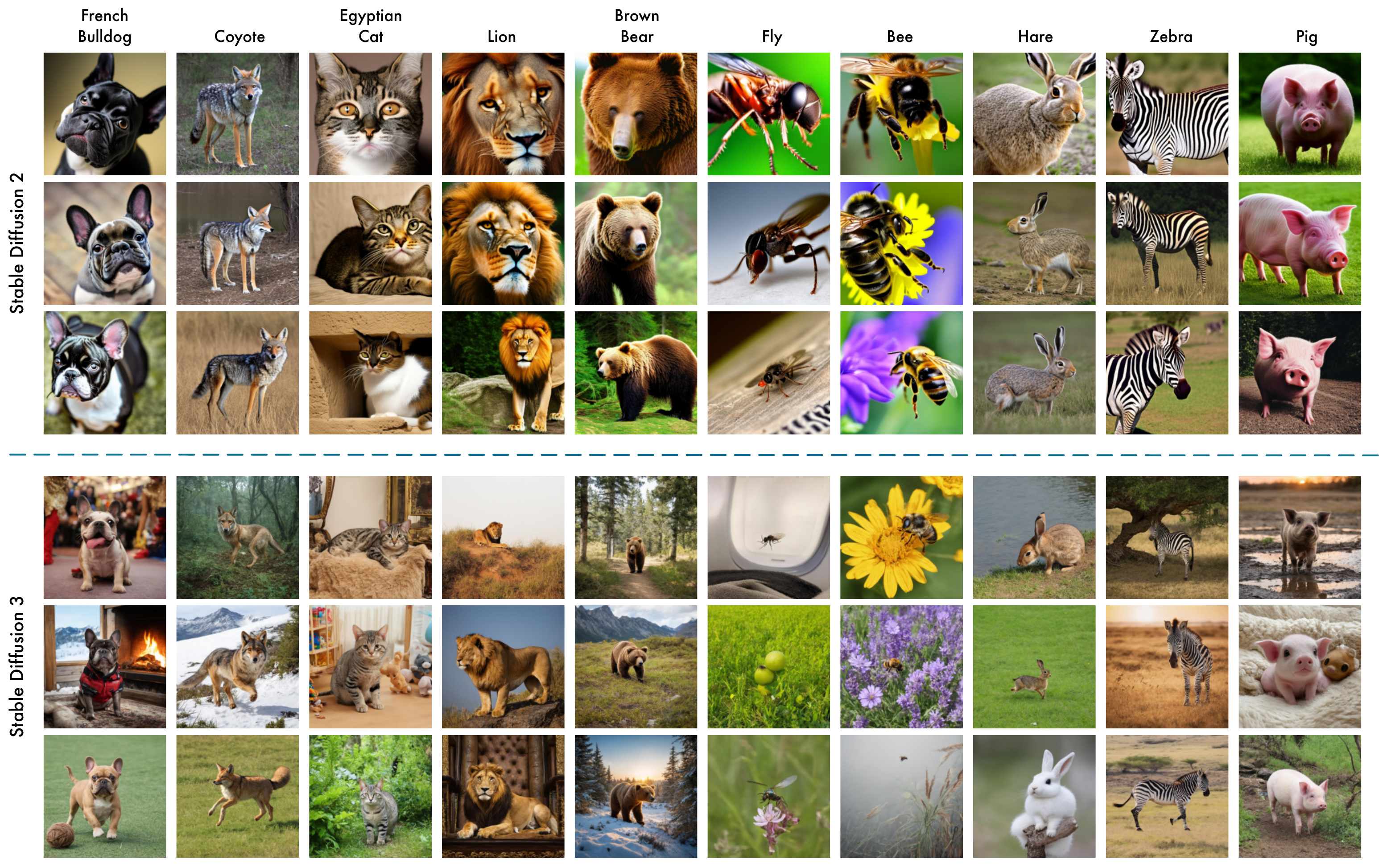}}
    \vspace{-5pt}
	\caption{Visualizations of the SD2-generated and SD3-generated synthetic dataset for ImageNet-10. SD2-generated images are often object-centric, focusing predominantly on the object's face. The backgrounds, shapes, and poses are usually uniform. In contrast, SD3-generated images present a more complete view of the objects, with diverse backgrounds and varied poses.}
	\label{fig:compare_quality}
\end{center}

\end{figure*}

\begin{figure*}[h]
\begin{center}
\centerline{\includegraphics[width=\textwidth]{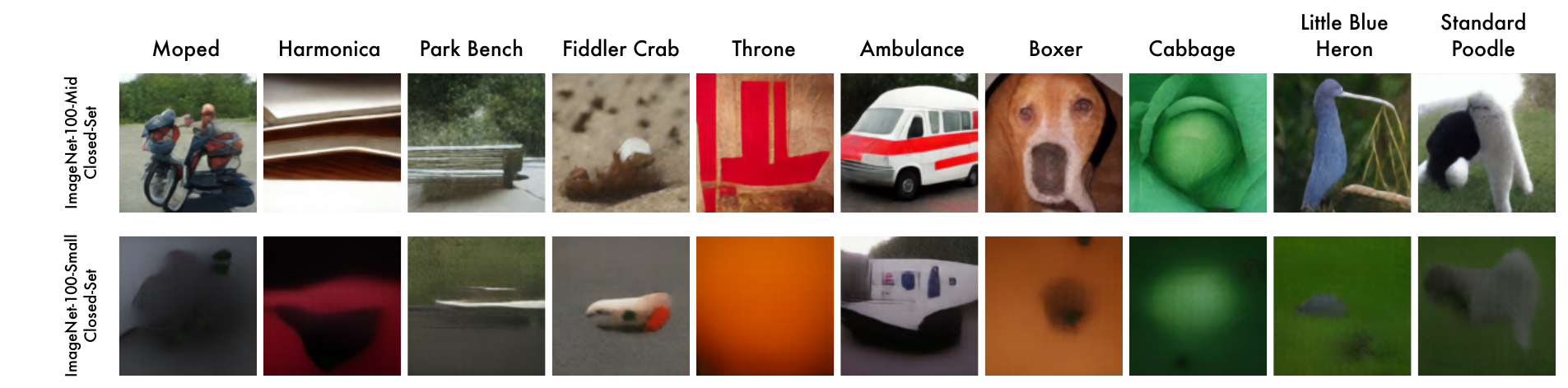}}
    \vspace{-5pt}
	\caption{Visualizations of the DiT-generated synthetic dataset for ImageNet-100-Mid and ImageNet-100-Small. The visualization for the synthetic data of ImageNet-100-Large is in Fig. 6 in the main text.}
	\label{fig:imgnet_100_mid_small}
\end{center}

\end{figure*}

\begin{figure*}[!h]
\begin{center}
\centerline{\includegraphics[width=\textwidth]{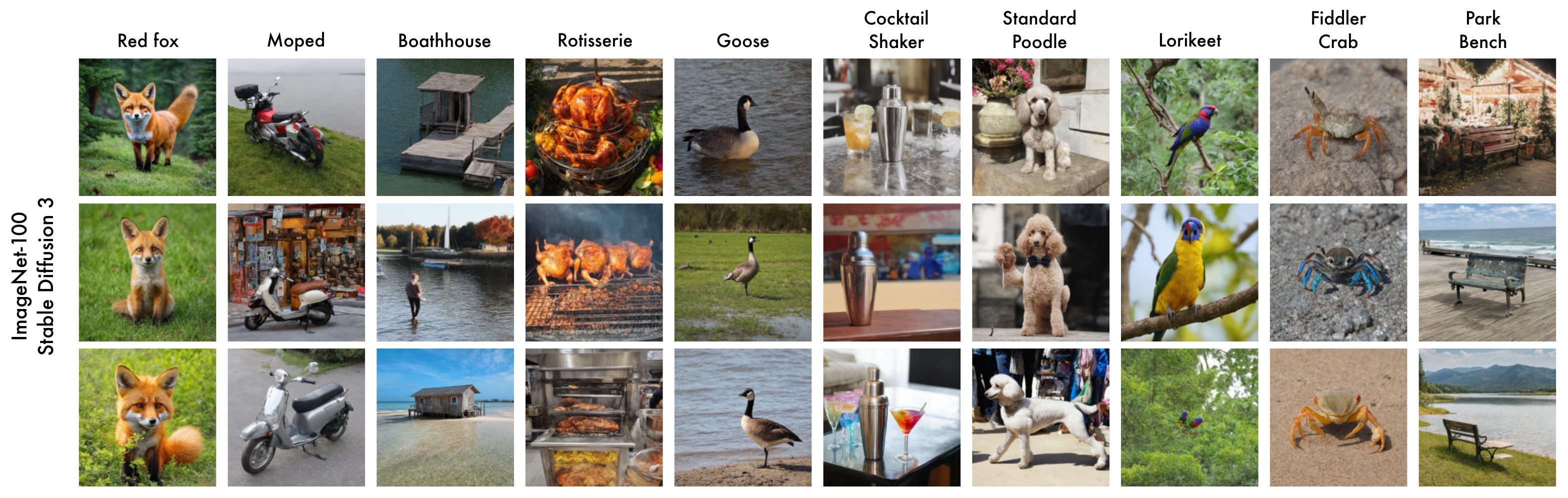}}
    \vspace{-5pt}
	\caption{Visualizations of the SD3-generated synthetic dataset for ImageNet-100.}
	\label{fig:appendix_vis_imagenet-100}
\end{center}

\end{figure*}

We visualize the synthetic datasets for ImageNet-10 and ImageNet-100. For ImageNet-10, we generate open-set synthetic images using Stable Diffusion 2 and 3. For ImageNet-100, closed-set generative images are produced using DiT, while open-set generative images are generated with Stable Diffusion 3. To visualize ImageNet-100, we randomly select ten classes from the entire dataset. All example images are randomly sampled from their respective classes without manual curation. The visualizations are presented in Fig. \ref{fig:compare_quality}, Fig. \ref{fig:imgnet_100_mid_small}, and Fig. \ref{fig:appendix_vis_imagenet-100}.

\section{Full Experimental Results}
We present all the experimental results we have used in this study. The results for real, closed-set, and open-set generative data augmentation on CIFAR-10 are presented in Tab. \ref{tab:appendix_cifar_results_real}, Tab. \ref{tab:appendix_cifar_results_edm}
 and Tab. \ref{tab:appendix_cifar_cifake_results}, respectively. The results for real and closed-set generative data augmentation on BloodMNIST are presented in Tab. \ref{tab:appendix_bloodmnist_results_real} and Tab. \ref{tab:appendix_bloodmnist_edm_results}, respectively. The results for real, closed-set, and open-set generative data augmentation on ImageNet-100 under the supervised classification setting are presented in Tab. \ref{tab:appendix_imagenet_100_real_results}, Tab. \ref{tab:appendix_imagenet_100_DiT_results},
 and Tab. \ref{tab:appendix_imagenet_100_C_results}, respectively. The results for generalization studies on ViT-B/32 are presented in Tab. \ref{tab:appendix_imagenet_100_real_results_vit}, Tab. \ref{tab:appendix_imagenet_100_DiT_results_vit},
 and Tab. \ref{tab:appendix_imagenet_100_C_results_vit}, respectively. The results for open-set generative data augmentation on ImageNet-10 under the supervised classification setting are presented in Tab. \ref{tab:appendix_imagenet_10_AB_results}. The results for open-set generative data augmentation on ImageNet-10 and ImageNet-100 under the zero-shot classification setting are presented in Tab. \ref{tab:appendix_zero-shot_imagenet_ABC_results}.

\clearpage

\begin{table*}[h]

  \vskip 0.05in 
  \centering
   \caption{Experimental results on CIFAR-10 with real data augmentation.}\label{tab:appendix_cifar_results_real}
  \begin{threeparttable}
  \begin{small}
  \renewcommand{\multirowsetup}{\centering}
  \setlength{\tabcolsep}{8pt}
  \begin{tabular}{c|ccc|c|c}
    \toprule
    \multirow{2}{*}
    {Training Dataset} & 
    \multicolumn{3}{c|}{{Data Amount}} &
    \multirow{2}{*}{\scalebox{1.0}{Base:Added}} &
    \scalebox{1.0}{Acc} \\
    \cmidrule(lr){2-4} \cmidrule(lr){6-6} &   
     \scalebox{1.0}{Base Real} & \scalebox{1.0}{Added Real}& Total & &\scalebox{1.0}{Top-1}\\
    \toprule
    \multirow{12}{*}{\scalebox{1.0}{CIFAR-10}} & \multirow{4}{*}{500} & 0 & 500 & base only & \scalebox{1.0}{33.17} \\
     
       & & 500 & 1,000 & 1:1 &{{\scalebox{1.0}{40.83 \textcolor{mygreen}{\text{\scriptsize (+7.66)}}}}} \\
        
        & & 1,000 & 1,500 & 1:2 &{{\scalebox{1.0}{48.81 \textcolor{mygreen}{\text{\scriptsize (+15.64)}}}}} \\
         
         & & 1,500 & 2,000 & 1:3 &{{\scalebox{1.0}{\textbf{54.30 \textcolor{mygreen}{\text{\scriptsize (+21.13)}}}}}} \\

  \cmidrule(lr){2-6}
    & \multirow{4}{*}{\scalebox{1.0}{5,000}} & 0 & 5,000 & base only &\scalebox{1.0}{74.20} \\
    & & 5,000& 10,000 & 1:1 & \scalebox{1.0}{83.56 \textcolor{mygreen}{\text{\scriptsize (+9.36)}}} \\
    & & 10,000& 15,000 & 1:2 & \scalebox{1.0}{87.29 \textcolor{mygreen}{\text{\scriptsize (+13.09)}}} \\
    & & 15,000& 20,000 & 1:3 &\scalebox{1.0}{\textbf{89.22 \textcolor{mygreen}{\text{\scriptsize (+15.02)}}}} \\

    \cmidrule(lr){2-6}
    & \multirow{4}{*}{25,000} & 0 & 25,000 & base only & \scalebox{1.0}{89.87} \\
     
       & & 10,000 & 35,000 & 1:0.4 &{{\scalebox{1.0}{91.85 \textcolor{mygreen}{\text{\scriptsize (+1.98)}}}}} \\
         & & 15,000 & 45,000 & 1:0.8 &{{\scalebox{1.0}{93.04 \textcolor{mygreen}{\text{\scriptsize (+3.17)}}}}} \\
  & & 25,000 & 50,000 & 1:1 &{{\scalebox{1.0}{\textbf{93.42 \textcolor{mygreen}{\text{\scriptsize (+3.55)}}}}}} \\

    \bottomrule
  \end{tabular}
    \end{small}
  \end{threeparttable}

\end{table*}

\begin{table*}[h]

  \vskip 0.05in 
  \centering
   \caption{Experimental results on CIFAR-10 using synthetic data produced by EDM for closed-set generative data augmentation.}\label{tab:appendix_cifar_results_edm}
  \begin{threeparttable}
  \begin{small}
  \renewcommand{\multirowsetup}{\centering}
  \setlength{\tabcolsep}{10pt}
  \begin{tabular}{cc|ccc|c|c}
    \toprule
    \multicolumn{2}{c|}{Training Dataset} & 
    \multicolumn{3}{c|}{{Data Amount}} &
    \multirow{2}{*}{\scalebox{1.0}{Real:Syn}} &
    \scalebox{1.0}{Acc} \\
    \cmidrule(lr){1-2} \cmidrule(lr){3-5} \cmidrule(lr){7-7}    
     \scalebox{1.0}{Real} & \scalebox{1.0}{Syn} & \scalebox{1.0}{Base Real} & \scalebox{1.0}{Added Syn}& Total & &\scalebox{1.0}{Top-1}\\
    \toprule
    \multirow{25}{*}{\scalebox{1.0}{CIFAR-10}}
    & \multirow{25}{*}{\scalebox{1.0}{EDM}} & \multirow{6}{*}{500} & 0 & 500 & real only & \scalebox{1.0}{33.17} \\
    
       & &  & 500 & 1,000 & 1:1 &{{\scalebox{1.0}{37.23 \textcolor{mygreen}{\text{\scriptsize (+4.06)}}}}} \\
        
        & &  & 1,000 & 1,500 & 1:2 &{{\scalebox{1.0}{40.45 \textcolor{mygreen}{\text{\scriptsize (+7.28)}}}}} \\
         
         & &  & 1,500 & 2,000 & 1:3 &{{\scalebox{1.0}{41.27 \textcolor{mygreen}{\text{\scriptsize (+8.10)}}}}} \\
  
  & &  & 2,000 & 2,500 & 1:4 &{{\scalebox{1.0}{43.38 \textcolor{mygreen}{\text{\scriptsize (+10.21)}}}}} \\
  & &  & 2,500 & 3,000 & 1:5 &{{\scalebox{1.0}{\textbf{44.58 \textcolor{mygreen}{\text{\scriptsize (+11.41)}}}}}} \\
  \cmidrule(lr){3-7}
    & & \multirow{6}{*}{\scalebox{1.0}{5,000}} & 0 & 5,000 & real only &\scalebox{1.0}{74.20} \\
    & &  & 5,000& 10,000 & 1:1 & \scalebox{1.0}{81.51 \textcolor{mygreen}{\text{\scriptsize (+7.31)}}} \\
    & &  & 10,000& 15,000 & 1:2 & \scalebox{1.0}{83.66 \textcolor{mygreen}{\text{\scriptsize (+9.46)}}} \\
    & &  & 15,000& 20,000 & 1:3 &\scalebox{1.0}{84.67 \textcolor{mygreen}{\text{\scriptsize (+10.47)}}} \\
    & &  & 20,000& 25,000  & 1:4 &\scalebox{1.0}{85.59 \textcolor{mygreen}{\text{\scriptsize (+11.39)}}} \\
    & &  & 25,000& 30,000  & 1:5 &\scalebox{1.0}{\textbf{86.17 \textcolor{mygreen}{\text{\scriptsize (+11.97)}}}} \\
   
    \cmidrule(lr){3-7}
    & & \multirow{6}{*}{25,000} & 0 & 25,000 & real only & \scalebox{1.0}{89.87} \\
     
       & &  & 25,000 & 50,000 & 1:1 &{{\scalebox{1.0}{92.33 \textcolor{mygreen}{\text{\scriptsize (+2.46)}}}}} \\
        
        & &  & 50,000 & 75,000 & 1:2 &{{\scalebox{1.0}{93.37 \textcolor{mygreen}{\text{\scriptsize (+3.50)}}}}} \\
         
         & &  & 75,000 & 100,000 & 1:3 &{{\scalebox{1.0}{94.07 \textcolor{mygreen}{\text{\scriptsize (+4.20)}}}}} \\
  
  & &  & 100,000 & 125,000 & 1:4 &{{\scalebox{1.0}{94.36 \textcolor{mygreen}{\text{\scriptsize (+4.49)}}}}} \\
  & &  & 125,000 & 150,000 & 1:5 &{{\scalebox{1.0}{\textbf{94.71 \textcolor{mygreen}{\text{\scriptsize (+4.84)}}}}}} \\
    
    \cmidrule(lr){3-7}
    & & \multirow{6}{*}{\scalebox{1.0}{50,000}} & 0 & 50,000  & real only & \scalebox{1.0}{93.42} \\
    & &  & 50,000& 100,000  & 1:1 & \scalebox{1.0}{95.35 \textcolor{mygreen}{\text{\scriptsize (+1.93)}}} \\
    
    & &  & 100,000 & 150,000 & 1:2 & \scalebox{1.0}{95.72 \textcolor{mygreen}{\text{\scriptsize (+2.30)}}} \\
     & &  & 150,000 & 200,000 & 1:3 &{{\scalebox{1.0}{96.06 \textcolor{mygreen}{\text{\scriptsize (+2.64)}}}}} \\
      & &  & 200,000 & 250,000 & 1:4 &{{\scalebox{1.0}{96.32 \textcolor{mygreen}{\text{\scriptsize (+2.90)}}}}} \\
       & &  & 250,000 & 300,000 & 1:5 & {{\scalebox{1.0}{\textbf{96.42 \textcolor{mygreen}{\text{\scriptsize (+3.00)}}}}}} \\

    \bottomrule
  \end{tabular}
    \end{small}
  \end{threeparttable}
   
\end{table*}

\begin{table*}[hbp]

  \vskip 0.05in 
  \centering
  \caption{Experimental results on CIFAR-10 using the CIFAKE synthetic dataset for open-set generative data augmentation.}\label{tab:appendix_cifar_cifake_results}
  \begin{threeparttable}
  \begin{small}
  \renewcommand{\multirowsetup}{\centering}
  \setlength{\tabcolsep}{10pt}
  \begin{tabular}{cc|ccc|c|c}
    \toprule
    \multicolumn{2}{c|}{Training Dataset} & 
    \multicolumn{3}{c|}{{Data Amount}} &
    \multirow{2}{*}{\scalebox{1.0}{Real:Syn}} &
    \scalebox{1.0}{Acc} \\
    \cmidrule(lr){1-2} \cmidrule(lr){3-5} \cmidrule(lr){7-7}    
     \scalebox{1.0}{Real} & \scalebox{1.0}{Syn} & \scalebox{1.0}{Base Real} & \scalebox{1.0}{Added Syn}& Total & & \scalebox{1.0}{Top-1}\\
    \toprule
    \multirow{20}{*}{\scalebox{1.0}{CIFAR-10}}
    & \multirow{20}{*}{\scalebox{1.0}{CIFAKE}} & \multirow{6}{*}{\scalebox{1.0}{500}} & 0 & 500 & real only & \scalebox{1.0}{33.17} \\
    & &  & 500& 1,000 & 1:1 & \scalebox{1.0}{40.53 \textcolor{mygreen}{\text{\scriptsize (+7.36)}}} \\
    & &  & 1,000& 1,500 & 1:2 & \scalebox{1.0}{43.01 \textcolor{mygreen}{\text{\scriptsize (+9.84)}}}\\
    & &  & 1,500& 2,000 & 1:3 & \scalebox{1.0}{45.11 \textcolor{mygreen}{\text{\scriptsize (+11.94)}}} \\
    & &  & 2,000& 2,500 & 1:4 & \scalebox{1.0}{{49.32 \textcolor{mygreen}{\text{\scriptsize (+16.15)}}}} \\
    & &  & 2,500& 3,000 & 1:5 & \scalebox{1.0}{\textbf{51.81 \textcolor{mygreen}{\text{\scriptsize (+18.64)}}}} \\
    \cmidrule(lr){3-7}
    &  & \multirow{6}{*}{\scalebox{1.0}{5,000}} & 0 & 5,000 & real only & \scalebox{1.0}{74.20} \\
    & &  & 5,000& 10,000 & 1:1 & \scalebox{1.0}{79.01 \textcolor{mygreen}{\text{\scriptsize (+4.81)}}} \\
    & &  & 10,000& 15,000 & 1:2 & \scalebox{1.0}{80.55 \textcolor{mygreen}{\text{\scriptsize (+6.35)}}}\\
    & &  & 15,000& 20,000 & 1:3 & \scalebox{1.0}{81.49 \textcolor{mygreen}{\text{\scriptsize (+7.29)}}} \\
    & &  & 20,000& 25,000 & 1:4 & \scalebox{1.0}{{82.30 \textcolor{mygreen}{\text{\scriptsize (+8.10)}}}} \\
    & &  & 25,000& 30,000 & 1:5 & \scalebox{1.0}{\textbf{82.61 \textcolor{mygreen}{\text{\scriptsize (+8.41)}}}} \\
    
    \cmidrule(lr){3-7}
    & & \multirow{5}{*}{25,000} & 0 & 25,000 & real only & \scalebox{1.0}{89.87} \\
     
     &&&10,000&35,000&1:0.4& {{\scalebox{1.0}{90.81 \textcolor{mygreen}{\text{\scriptsize (+0.94)}}}}} \\
     & & & 20,000 & 45,000 & 1:0.8 &{{\scalebox{1.0}{90.95 \textcolor{mygreen}{\text{\scriptsize (+1.08)}}}}} \\
       & &  & 25,000 & 50,000 & 1:1 &{{\scalebox{1.0}{90.96 \textcolor{mygreen}{\text{\scriptsize (+1.09)}}}}} \\
        
        & &  & 50,000 & 75,000 & 1:2 &{{\scalebox{1.0}{\textbf{91.66 \textcolor{mygreen}{\text{\scriptsize (+1.79)}}}}}} \\
   
    \cmidrule(lr){3-7}
    & & \multirow{4}{*}{\scalebox{1.0}{50,000}} & 0 & 50,000 & real only & \scalebox{1.0}{93.42} \\
    
    & &  & 20,000& 70,000 & 1:0.4 & \scalebox{1.0}{94.05 \textcolor{mygreen}{\text{\scriptsize (+0.63)}}} \\
    
    & &  & 40,000& 90,000 & 1:0.8 & \scalebox{1.0}{93.96 \textcolor{mygreen}{\text{\scriptsize (+0.54)}}} \\
    & &  & 50,000& 100,000 & 1:1 & \scalebox{1.0}{\textbf{94.11 \textcolor{mygreen}{\text{\scriptsize (+0.69)}}}} \\

    \bottomrule
  \end{tabular}
    \end{small}
  \end{threeparttable}
    
\end{table*}

\begin{table*}[h]
  \vskip 0.05in 
  \centering
      \caption{Experimental results on BloodMNIST with real data augmentation.}\label{tab:appendix_bloodmnist_results_real}
  \begin{threeparttable}
  \begin{small}
  \renewcommand{\multirowsetup}{\centering}
  \setlength{\tabcolsep}{10pt}
  \begin{tabular}{c|ccc|c|c}
    \toprule
    \multirow{2}{*}
    {Training Dataset} & 
    \multicolumn{3}{c|}{{Data Amount}} &
    \multirow{2}{*}{\scalebox{1.0}{Base:Added}} &
    \scalebox{1.0}{Acc} \\
    \cmidrule(lr){2-4} \cmidrule(lr){6-6} &   
     \scalebox{1.0}{Base Real} & \scalebox{1.0}{Added Real}& Total & &\scalebox{1.0}{Top-1}\\
    \toprule
    \multirow{7}{*}{\scalebox{1.0}{BloodMNIST}} & \multirow{4}{*}{1,200} & 0 & 1,200 & base only & \scalebox{1.0}{91.80} \\
       & & 1,200 & 2,400 & 1:1 &{{\scalebox{1.0}{94.55 \textcolor{mygreen}{\text{\scriptsize (+2.75)}}}}} \\
        & & 2,400 & 3,600 & 1:2 &{{\scalebox{1.0}{96.46 \textcolor{mygreen}{\text{\scriptsize (+4.66)}}}}} \\
         & & 3,600 & 4,800 & 1:3 &{{\scalebox{1.0}{\textbf{96.77 \textcolor{mygreen}{\text{\scriptsize (+4.97)}}}}}} \\

    \cmidrule(lr){2-6}
    & \multirow{3}{*}{6,000} & 0 & 6,000 & base only & \scalebox{1.0}{97.05} \\
       & & 3,000 & 9,000 & 1:0.5 &{{\scalebox{1.0}{97.55 \textcolor{mygreen}{\text{\scriptsize (+0.50)}}}}} \\
         & & 6,000 & 12,000 & 1:1 &{{\scalebox{1.0}{\textbf{97.83 \textcolor{mygreen}{\text{\scriptsize (+0.78)}}}}}} \\
    \bottomrule
  \end{tabular}
    \end{small}
  \end{threeparttable}

\end{table*}

\begin{table*}
  \vskip 0.05in 
  \centering
  \caption{Experimental results on BloodMNIST using EDM for closed-set generative data augmentation.}\label{tab:appendix_bloodmnist_edm_results}
  \begin{threeparttable}
  \begin{small}
  \renewcommand{\multirowsetup}{\centering}
  \setlength{\tabcolsep}{10pt}
  \begin{tabular}{cc|ccc|c|c}
    \toprule
    \multicolumn{2}{c|}{Training Dataset} & 
    \multicolumn{3}{c|}{{Data Amount}} &
    \multirow{2}{*}{\scalebox{1.0}{Real:Syn}} &
    \scalebox{1.0}{Acc} \\
    \cmidrule(lr){1-2} \cmidrule(lr){3-5} \cmidrule(lr){7-7}    
     \scalebox{1.0}{Real} & \scalebox{1.0}{Syn} & \scalebox{1.0}{Base Real} & \scalebox{1.0}{Added Syn}& Total & & \scalebox{1.0}{Top-1}\\
    \toprule
    \multirow{18}{*}{\scalebox{1.0}{BloodMNIST}}
    & \multirow{18}{*}{\scalebox{1.0}{EDM}} & \multirow{6}{*}{\scalebox{1.0}{1,200}} & 0 & 1,200 & real only & \scalebox{1.0}{91.80} \\

    & & & 1,200 & 2,400 & 1:1 & \scalebox{1.0}{94.21  \textcolor{mygreen}{\text{\scriptsize (+2.41)}}} \\

     & & & 2,400 & 3,600 & 1:2 & \scalebox{1.0}{94.60  \textcolor{mygreen}{\text{\scriptsize (+2.80)}}} \\

     & & & 3,600 & 4,800 & 1:3 & \scalebox{1.0}{94.72  \textcolor{mygreen}{\text{\scriptsize (+2.92)}}} \\

     & & & 4,800 & 6,000 & 1:4 & \scalebox{1.0}{\textbf{95.32  \textcolor{mygreen}{\text{\scriptsize (+3.52)}}}} \\

     & & & 6,000 & 7,200 & 1:5 & \scalebox{1.0}{95.19  \textcolor{mygreen}{\text{\scriptsize (+3.39)}}} \\

     \cmidrule{3-7}

     & & \multirow{6}{*}{\scalebox{1.0}{6,000}} & 0 & 6,000 & real only & \scalebox{1.0}{97.05} \\

     & & & 6,000 & 12,000 & 1:1 & \scalebox{1.0}{97.62  \textcolor{mygreen}{\text{\scriptsize (+0.57)}}} \\

     & & & 12,000 & 18,000 & 1:2 & \scalebox{1.0}{97.95  \textcolor{mygreen}{\text{\scriptsize (+0.90)}}} \\

     & & & 18,000 & 24,000 & 1:3 & \scalebox{1.0}{98.06  \textcolor{mygreen}{\text{\scriptsize (+1.01)}}} \\

     & & & 24,000 & 30,000 & 1:4 & \scalebox{1.0}{98.12  \textcolor{mygreen}{\text{\scriptsize (+1.07)}}} \\

     & & & 30,000 & 36,000 &  1:5 & \scalebox{1.0}{\textbf{98.28 \textcolor{mygreen}{\text{\scriptsize (+1.23)}}}} \\

     \cmidrule{3-7}

     & & \multirow{6}{*}{\scalebox{1.0}{12,000}} & 0 & 12,000 & real only & \scalebox{1.0}{97.83} \\

     & & & 12,000 & 24,000 & 1:1 & \scalebox{1.0}{98.27  \textcolor{mygreen}{\text{\scriptsize (+0.44)}}} \\

     & & & 24,000 & 36,000 & 1:2 & \scalebox{1.0}{98.43  \textcolor{mygreen}{\text{\scriptsize (+0.60)}}} \\

     & & & 36,000 & 48,000 & 1:3 & \scalebox{1.0}{98.48  \textcolor{mygreen}{\text{\scriptsize (+0.65)}}} \\

     & & & 48,000 & 60,000 & 1:4 & \scalebox{1.0}{\textbf{98.56  \textcolor{mygreen}{\text{\scriptsize (+0.73)}}}} \\

     & & & 60,000 & 72,000 & 1:5 & \scalebox{1.0}{98.48 \textcolor{mygreen}{\text{\scriptsize (+0.65)}}} \\
     
\bottomrule
  \end{tabular}
    \end{small}
  \end{threeparttable}
    
\end{table*}

\begin{table*}
  \vskip 0.05in
   \renewcommand{\arraystretch}{1.0} 
  \centering
   \caption{Experimental results of supervised image classification with real data augmentation on ImageNet-100.}\label{tab:appendix_imagenet_100_real_results}
  \begin{threeparttable}
  \begin{small}
  \renewcommand{\multirowsetup}{\centering}
  \setlength{\tabcolsep}{6pt}
  \begin{tabular}{c|ccc|c|cc}
    \toprule
    \multirow{2}{*}{\scalebox{1.0}{Training Dataset}} & 
    \multicolumn{3}{c|}{\rotatebox{0}{\scalebox{1.0}{Data Amount}}} &
    \multirow{2}{*}{\scalebox{1.0}{Base:Added}} &
    \multicolumn{2}{c}{\scalebox{1.0}{Acc}} \\
    \cmidrule(lr){2-4} \cmidrule(lr){6-7}    
    & \scalebox{1.0}{Base Real} & \scalebox{1.0}{Added Real}& Total & & \scalebox{1.0}{Top-1} &\scalebox{1.0}{Top-5}\\
    \toprule
    \multirow{10}{*}{\scalebox{1.0}{ImageNet-100}} & \multirow{6}{*}{\scalebox{1.0}{6,500}} & 0 & 6,500 & base only & \scalebox{1.0}{26.07} & \scalebox{1.0}{49.93} \\
    & &  650& 7,150 & 1:0.1 & \scalebox{1.0}{30.37 \textcolor{mygreen}{\text{\scriptsize (+4.30)}}} & \scalebox{1.0}{54.47 \textcolor{mygreen}{\text{\scriptsize (+4.54)}}} \\
    
    &  & 3,250 & 9,750 &1:0.5 & \scalebox{1.0}{37.13  \textcolor{mygreen}{\text{\scriptsize (+11.06)}}} & \scalebox{1.0}{61.63  \textcolor{mygreen}{\text{\scriptsize (+11.70)}}} \\
     &  & 6,500 & 13,000 & 1:1& {{\scalebox{1.0}{43.21 \textcolor{mygreen}{\text{\scriptsize (+17.14)}}}}} & \scalebox{1.0}{67.60  \textcolor{mygreen}{\text{\scriptsize (+17.67)}}}  \\
     &  & 26,000 & 32,500 & 1:4& {{\scalebox{1.0}{67.40 \textcolor{mygreen}{\text{\scriptsize (+41.33)}}}}} & {{\scalebox{1.0}{86.19 \textcolor{mygreen}{\text{\scriptsize (+36.26)}}}}} \\
      &  & 65,000 & 71,500 & 1:10& \textbf{{{\scalebox{1.0}{79.93 \textcolor{mygreen}{\text{\scriptsize (+53.86)}}}}}} & \textbf{{{\scalebox{1.0}{93.63 \textcolor{mygreen}{\text{\scriptsize (+43.70)}}}}}} \\

    \cmidrule(lr){2-7}
    & \multirow{5}{*}{\scalebox{1.0}{26,000}} & 0 & 26,000 & base only & \scalebox{1.0}{62.78} & \scalebox{1.0}{82.90} \\
    &  & 2,600& 28,600 & 1:0.1 & \scalebox{1.0}{65.27 \textcolor{mygreen}{\text{\scriptsize (+2.49)}}} & \scalebox{1.0}{84.93 \textcolor{mygreen}{\text{\scriptsize (+2.03)}}} \\
    
    &  & 13,000 & 39,000 &1:0.5 & \scalebox{1.0}{70.73  \textcolor{mygreen}{\text{\scriptsize (+7.95)}}} & \scalebox{1.0}{88.53 \textcolor{mygreen}{\text{\scriptsize (+5.63)}}} \\
     &  & 26,000 & 52,000 & 1:1& {{\scalebox{1.0}{75.52 \textcolor{mygreen}{\text{\scriptsize (+12.74)}}}}} & \scalebox{1.0}{91.06 \textcolor{mygreen}{\text{\scriptsize (+8.15)}}} \\
      &  & 104,000 & 130,000 & 1:4& \textbf{{{\scalebox{1.0}{85.41 \textcolor{mygreen}{\text{\scriptsize (+22.63)}}}}}} & \textbf{\scalebox{1.0}{96.39 \textcolor{mygreen}{\text{\scriptsize (+13.49)}}}} \\
    \bottomrule
  \end{tabular}
    \end{small}
  \end{threeparttable}
\end{table*}

\begin{table*}
  \vskip 0.05in
   \renewcommand{\arraystretch}{1.0} 
  \centering
  \caption{Experimental results of supervised image classification with closed-set generative data augmentation on ImageNet-100.}\label{tab:appendix_imagenet_100_DiT_results}
  \begin{threeparttable}
  \begin{small}
  \renewcommand{\multirowsetup}{\centering}
  \setlength{\tabcolsep}{3pt}
  \begin{tabular}{cc|ccc|c|cc}
    \toprule
    \multicolumn{2}{c|}{Training Dataset} & 
    \multicolumn{3}{c|}{\rotatebox{0}{\scalebox{1.0}{Data Amount}}} &
    \multirow{2}{*}{\scalebox{1.0}{Real:Syn }} &
    \multicolumn{2}{c}{\scalebox{1.0}{Acc}} \\
    \cmidrule(lr){1-2} \cmidrule(lr){3-5} \cmidrule(lr){7-8}    
     \scalebox{1.0}{Real} & \scalebox{1.0}{Syn} & \scalebox{1.0}{Base Real} & \scalebox{1.0}{Added Syn}& Total & & \scalebox{1.0}{Top-1} &\scalebox{1.0}{Top-5}\\
    \toprule
    \multirow{16}{*}{\scalebox{1.0}{ImageNet-100}}
    & \multirow{16}{*}{\scalebox{1.0}{ImageNet-100-DiT}} & \multirow{5}{*}{\scalebox{1.0}{6,500}} & 0 & 6,500 & real only & \scalebox{1.0}{26.07} & \scalebox{1.0}{49.93} \\
    & &  & 650& 7,150 & 1:0.1 & \scalebox{1.0}{31.47 \textcolor{mygreen}{\text{\scriptsize (+5.40)}}} & \scalebox{1.0}{55.14 \textcolor{mygreen}{\text{\scriptsize (+5.21)}}} \\
    
    & &  & 3,250 & 9,750 &1:0.5 & \scalebox{1.0}{31.75  \textcolor{mygreen}{\text{\scriptsize (+5.68)}}} & \scalebox{1.0}{56.36  \textcolor{mygreen}{\text{\scriptsize (+6.43)}}} \\
     & &  & 6,500 & 13,000 & 1:1& {{\scalebox{1.0}{34.36 \textcolor{mygreen}{\text{\scriptsize (+8.29)}}}}} & \scalebox{1.0}{58.67  \textcolor{mygreen}{\text{\scriptsize (+8.74)}}}  \\
     & &  & 26,000 & 32,500 & 1:4& {{\scalebox{1.0}{36.27 \textcolor{mygreen}{\text{\scriptsize (+10.20)}}}}} & \scalebox{1.0}{61.30  \textcolor{mygreen}{\text{\scriptsize (+11.37)}}}  \\
      & &  & 65,000 & 71,500 & 1:10& \textbf{{{\scalebox{1.0}{36.61 \textcolor{mygreen}{\text{\scriptsize (+10.54)}}}}}} & \textbf{{{\scalebox{1.0}{62.79 \textcolor{mygreen}{\text{\scriptsize (+12.86)}}}}}} \\

    \cmidrule(lr){3-8}
    & & \multirow{5}{*}{\scalebox{1.0}{26,000}} & 0 & 26,000 & real only & \scalebox{1.0}{62.78} & \scalebox{1.0}{82.90} \\
    & &  & 2,600& 28,600 & 1:0.1 & \scalebox{1.0}{65.97 \textcolor{mygreen}{\text{\scriptsize (+3.19)}}} & \scalebox{1.0}{85.59 \textcolor{mygreen}{\text{\scriptsize (+2.69)}}} \\
  
    & &  & 13,000 & 39,000 &1:0.5 & \scalebox{1.0}{67.35  \textcolor{mygreen}{\text{\scriptsize (+4.57)}}} & \scalebox{1.0}{86.48 \textcolor{mygreen}{\text{\scriptsize (+3.58)}}} \\
     & &  & 26,000 & 52,000 & 1:1& {{\scalebox{1.0}{66.96 \textcolor{mygreen}{\text{\scriptsize (+4.18)}}}}} & \scalebox{1.0}{86.53 \textcolor{mygreen}{\text{\scriptsize (+3.63)}}} \\
      & &  & 104,000 & 130,000 & 1:4& {{\scalebox{1.0}{69.53 \textcolor{mygreen}{\text{\scriptsize (+6.75)}}}}} & \scalebox{1.0}{88.66 \textcolor{mygreen}{\text{\scriptsize (+5.76)}}} \\
      & &  & 130,000 & 156,000 & 1:5& \textbf{{{\scalebox{1.0}{70.07 \textcolor{mygreen}{\text{\scriptsize (+7.29)}}}}}} & \textbf{\scalebox{1.0}{89.29 \textcolor{mygreen}{\text{\scriptsize (+6.39)}}}} \\

    \cmidrule(lr){3-8}
    & & \multirow{4}{*}{\scalebox{1.0}{130,000}} & 0 & 130,000 & real only & \scalebox{1.0}{85.41} & \scalebox{1.0}{96.39} \\
    & &  & 13,000& 143,000 & 1:0.1 & \scalebox{1.0}{85.59 \textcolor{mygreen}{\text{\scriptsize (+0.18)}}} & \scalebox{1.0}{96.57 \textcolor{mygreen}{\text{\scriptsize (+0.18)}}} \\
   
    & &  & 65,000 & 195,000 &1:0.5 & \scalebox{1.0}{85.87  \textcolor{mygreen}{\text{\scriptsize (+0.46)}}} & \scalebox{1.0}{96.68  \textcolor{mygreen}{\text{\scriptsize (+0.29)}}} \\
     & &  & 130,000 & 260,000 & 1:1& \textbf{{{\scalebox{1.0}{86.12 \textcolor{mygreen}{\text{\scriptsize (+0.71)}}}}}} & \textbf{\scalebox{1.0}{96.71  \textcolor{mygreen}{\text{\scriptsize (+0.32)}}}} \\
    
    \bottomrule
  \end{tabular}
    \end{small}
  \end{threeparttable}
  
\end{table*}

\begin{table*}
  \vskip 0.05in
   \renewcommand{\arraystretch}{1.0} 
  \centering
  \caption{Experimental results of supervised image classification with open-set generative data augmentation on ImageNet-100.}\label{tab:appendix_imagenet_100_C_results}
  \begin{threeparttable}
  \begin{small}
  \renewcommand{\multirowsetup}{\centering}
  \setlength{\tabcolsep}{3pt}
  \begin{tabular}{cc|ccc|c|cc}
    \toprule
    \multicolumn{2}{c|}{Training Dataset} & 
    \multicolumn{3}{c|}{\rotatebox{0}{\scalebox{1.0}{Data Amount}}} &
    \multirow{2}{*}{\scalebox{1.0}{Real:Syn }} &
    \multicolumn{2}{c}{\scalebox{1.0}{Acc}} \\
    \cmidrule(lr){1-2} \cmidrule(lr){3-5} \cmidrule(lr){7-8}    
     \scalebox{1.0}{Real} & \scalebox{1.0}{Syn} & \scalebox{1.0}{Base Real} & \scalebox{1.0}{Added Syn}& Total & & \scalebox{1.0}{Top-1} &\scalebox{1.0}{Top-5}\\
    \toprule
    \multirow{16}{*}{\scalebox{1.0}{ImageNet-100}}
    & \multirow{16}{*}{\scalebox{1.0}{ImageNet-100-SD3}} & \multirow{5}{*}{\scalebox{1.0}{6,500}} & 0 & 6,500 & real only & \scalebox{1.0}{26.07} & \scalebox{1.0}{49.93} \\
    & &  & 650& 7,150 & 1:0.1 & \scalebox{1.0}{29.89 \textcolor{mygreen}{\text{\scriptsize (+3.82)}}} & \scalebox{1.0}{54.29 \textcolor{mygreen}{\text{\scriptsize (+4.36)}}} \\
    
    & &  & 3,250 & 9,750 &1:0.5 & \scalebox{1.0}{35.89  \textcolor{mygreen}{\text{\scriptsize (+9.82)}}} & \scalebox{1.0}{60.22  \textcolor{mygreen}{\text{\scriptsize (+10.29)}}} \\
     & &  & 6,500 & 13,000 & 1:1& {{\scalebox{1.0}{41.31 \textcolor{mygreen}{\text{\scriptsize (+15.24)}}}}} & \scalebox{1.0}{66.05  \textcolor{mygreen}{\text{\scriptsize (+16.12)}}}  \\
     & &  & 26,000 & 32,500 & 1:4& {{\scalebox{1.0}{57.47 \textcolor{mygreen}{\text{\scriptsize (+31.40)}}}}} & \scalebox{1.0}{80.33  \textcolor{mygreen}{\text{\scriptsize (+30.40)}}}  \\
      & &  & 65,000 & 71,500 & 1:10& \textbf{{{\scalebox{1.0}{64.78 \textcolor{mygreen}{\text{\scriptsize (+38.71)}}}}}} & \textbf{{{\scalebox{1.0}{85.83 \textcolor{mygreen}{\text{\scriptsize (+35.90)}}}}}} \\

    \cmidrule(lr){3-8}
    & & \multirow{5}{*}{\scalebox{1.0}{26,000}} & 0 & 26,000 & real only & \scalebox{1.0}{62.78} & \scalebox{1.0}{82.90} \\
    & &  & 2,600& 28,600 & 1:0.1 & \scalebox{1.0}{64.95 \textcolor{mygreen}{\text{\scriptsize (+2.17)}}} & \scalebox{1.0}{84.71 \textcolor{mygreen}{\text{\scriptsize (+1.81)}}} \\
   
    & &  & 13,000 & 39,000 &1:0.5 & \scalebox{1.0}{69.65  \textcolor{mygreen}{\text{\scriptsize (+6.87)}}} & \scalebox{1.0}{88.11 \textcolor{mygreen}{\text{\scriptsize (+5.21)}}} \\
     & &  & 26,000 & 52,000 & 1:1& {{\scalebox{1.0}{72.31 \textcolor{mygreen}{\text{\scriptsize (+9.53)}}}}} & \scalebox{1.0}{89.81 \textcolor{mygreen}{\text{\scriptsize (+6.91)}}} \\
      & &  & 104,000 & 130,000 & 1:4& {{\scalebox{1.0}{77.57 \textcolor{mygreen}{\text{\scriptsize (+14.79)}}}}} & \scalebox{1.0}{93.01 \textcolor{mygreen}{\text{\scriptsize (+10.11)}}} \\
      & &  & 130,000 & 156,000 & 1:5& \textbf{{{\scalebox{1.0}{78.17 \textcolor{mygreen}{\text{\scriptsize (+15.39)}}}}}} & \textbf{\scalebox{1.0}{93.49 \textcolor{mygreen}{\text{\scriptsize (+10.59)}}}} \\

    \cmidrule(lr){3-8}
    & & \multirow{4}{*}{\scalebox{1.0}{130,000}} & 0 & 130,000 & real only & \scalebox{1.0}{85.41} & \scalebox{1.0}{96.39} \\
    & &  & 13,000& 143,000 & 1:0.1 & \scalebox{1.0}{85.69 \textcolor{mygreen}{\text{\scriptsize (+0.28)}}} & \scalebox{1.0}{96.63 \textcolor{mygreen}{\text{\scriptsize (+0.24)}}} \\
    
    & &  & 65,000 & 195,000 &1:0.5 & \scalebox{1.0}{86.85  \textcolor{mygreen}{\text{\scriptsize (+1.44)}}} & \scalebox{1.0}{97.10  \textcolor{mygreen}{\text{\scriptsize (+0.71)}}} \\
     & &  & 130,000 & 260,000 & 1:1& \textbf{{{\scalebox{1.0}{86.91 \textcolor{mygreen}{\text{\scriptsize (+1.50)}}}}}} & \textbf{\scalebox{1.0}{97.29  \textcolor{mygreen}{\text{\scriptsize (+0.90)}}}} \\
    
    \bottomrule
  \end{tabular}
    \end{small}
  \end{threeparttable}
\end{table*}

\begin{table*}
  \vskip 0.05in
   \renewcommand{\arraystretch}{1.0} 
  \centering
   \caption{Experimental results of supervised image classification with real data augmentation on ImageNet-100 {\bf using ViT-B/32 as the classifier.}}\label{tab:appendix_imagenet_100_real_results_vit}
  \begin{threeparttable}
  \begin{small}
  \renewcommand{\multirowsetup}{\centering}
  \setlength{\tabcolsep}{6pt}
  \begin{tabular}{c|ccc|c|c}
    \toprule
    \multirow{2}{*}{\scalebox{1.0}{Training Dataset}} & 
    \multicolumn{3}{c|}{\rotatebox{0}{\scalebox{1.0}{Data Amount}}} &
    \multirow{2}{*}{\scalebox{1.0}{Base:Added}} &
    {\scalebox{1.0}{Acc}} \\
    \cmidrule(lr){2-4} \cmidrule(lr){6-6}    
    & \scalebox{1.0}{Base Real} & \scalebox{1.0}{Added Real}& Total & & \scalebox{1.0}{Top-1}\\
    \toprule
    \multirow{10}{*}{\scalebox{1.0}{ImageNet-100}} & \multirow{6}{*}{\scalebox{1.0}{6,500}} & 0 & 6,500 & base only & \scalebox{1.0}{15.57}\\
    & &  650& 7,150 & 1:0.1 & \scalebox{1.0}{15.97 \textcolor{mygreen}{\text{\scriptsize (+0.40)}}} \\
   
    &  & 3,250 & 9,750 &1:0.5 & \scalebox{1.0}{17.63  \textcolor{mygreen}{\text{\scriptsize (+2.06)}}} \\
     &  & 6,500 & 13,000 & 1:1& {{\scalebox{1.0}{20.61 \textcolor{mygreen}{\text{\scriptsize (+5.04)}}}}}\\
     &  & 26,000 & 32,500 & 1:4& {{\scalebox{1.0}{30.51 \textcolor{mygreen}{\text{\scriptsize (+14.94)}}}}}\\
      &  & 65,000 & 71,500 & 1:10& \textbf{{{\scalebox{1.0}{41.21 \textcolor{mygreen}{\text{\scriptsize (+25.64)}}}}}}\\

    \cmidrule(lr){2-6}
    & \multirow{5}{*}{\scalebox{1.0}{26,000}} & 0 & 26,000 & base only & \scalebox{1.0}{27.59}\\
    &  & 2,600& 28,600 & 1:0.1 & \scalebox{1.0}{28.73 \textcolor{mygreen}{\text{\scriptsize (+1.14)}}} \\
    
    &  & 13,000 & 39,000 &1:0.5 & \scalebox{1.0}{32.84  \textcolor{mygreen}{\text{\scriptsize (+5.25)}}} \\
     &  & 26,000 & 52,000 & 1:1& {{\scalebox{1.0}{36.94 \textcolor{mygreen}{\text{\scriptsize (+9.35)}}}}} \\
      &  & 104,000 & 130,000 & 1:4& \textbf{{{\scalebox{1.0}{48.91 \textcolor{mygreen}{\text{\scriptsize (+21.32)}}}}}}\\
    \bottomrule
  \end{tabular}
    \end{small}
  \end{threeparttable}
\end{table*}

\begin{table*}
  \vskip 0.05in
   \renewcommand{\arraystretch}{1.0} 
  \centering
  \caption{Experimental results of supervised image classification with closed-set generative data augmentation on ImageNet-100 {\bf using ViT-B/32 as the classifier.}}\label{tab:appendix_imagenet_100_DiT_results_vit}
  \begin{threeparttable}
  \begin{small}
  \renewcommand{\multirowsetup}{\centering}
  \setlength{\tabcolsep}{3pt}
  \begin{tabular}{cc|ccc|c|c}
    \toprule
    \multicolumn{2}{c|}{Training Dataset} & 
    \multicolumn{3}{c|}{\rotatebox{0}{\scalebox{1.0}{Data Amount}}} &
    \multirow{2}{*}{\scalebox{1.0}{Real:Syn }} & \scalebox{1.0}{Acc} \\
    \cmidrule(lr){1-2} \cmidrule(lr){3-5} \cmidrule(lr){7-7}    
     \scalebox{1.0}{Real} & \scalebox{1.0}{Syn} & \scalebox{1.0}{Base Real} & \scalebox{1.0}{Added Syn}& Total & & \scalebox{1.0}{Top-1} \\
    \toprule
    \multirow{16}{*}{\scalebox{1.0}{ImageNet-100}}
    & \multirow{16}{*}{\scalebox{1.0}{ImageNet-100-DiT}} & \multirow{5}{*}{\scalebox{1.0}{6,500}} & 0 & 6,500 & real only & \scalebox{1.0}{15.57} \\
    & &  & 650& 7,150 & 1:0.1 & \scalebox{1.0}{15.54 \textcolor{myred}{\text{\scriptsize (-0.03)}}} \\
   
    & &  & 3,250 & 9,750 &1:0.5 & \scalebox{1.0}{15.38  \textcolor{myred}{\text{\scriptsize (-0.19)}}} \\
     & &  & 6,500 & 13,000 & 1:1& {{\scalebox{1.0}{15.76 \textcolor{mygreen}{\text{\scriptsize (+0.19)}}}}}\\
     & &  & 26,000 & 32,500 & 1:4& {{\scalebox{1.0}{16.91 \textcolor{mygreen}{\text{\scriptsize (+1.34)}}}}} \\
      & &  & 65,000 & 71,500 & 1:10& \scalebox{1.0}{\bf 17.82  \textcolor{mygreen}{\text{\scriptsize (+2.25)}}}  \\

    \cmidrule(lr){3-7}
    & & \multirow{5}{*}{\scalebox{1.0}{26,000}} & 0 & 26,000 & real only & \scalebox{1.0}{27.59}\\
    & &  & 2,600& 28,600 & 1:0.1 & \scalebox{1.0}{28.28 \textcolor{mygreen}{\text{\scriptsize (+0.69)}}}\\
  
    & &  & 13,000 & 39,000 &1:0.5 & \scalebox{1.0}{29.45  \textcolor{mygreen}{\text{\scriptsize (+1.86)}}}  \\
     & &  & 26,000 & 52,000 & 1:1& {{\scalebox{1.0}{30.31 \textcolor{mygreen}{\text{\scriptsize (+2.72)}}}}} \\
      & &  & 104,000 & 130,000 & 1:4& \textbf{{{\scalebox{1.0}{33.07 \textcolor{mygreen}{\text{\scriptsize (+5.48)}}}}}}\\
      & &  & 130,000 & 156,000 & 1:5& \scalebox{1.0}{32.78 \textcolor{mygreen}{\text{\scriptsize (+5.19)}}} \\

    \cmidrule(lr){3-7}
    & & \multirow{4}{*}{\scalebox{1.0}{130,000}} & 0 & 130,000 & real only & \scalebox{1.0}{48.91} \\
    & &  & 13,000& 143,000 & 1:0.1 & \scalebox{1.0}{49.43 \textcolor{mygreen}{\text{\scriptsize (+0.52)}}}  \\
    
    & &  & 65,000 & 195,000 &1:0.5 & \scalebox{1.0}{50.47  \textcolor{mygreen}{\text{\scriptsize (+1.56)}}} \\
     & &  & 130,000 & 260,000 & 1:1& \textbf{{{\scalebox{1.0}{50.96 \textcolor{mygreen}{\text{\scriptsize (+2.05)}}}}}}\\
    
    \bottomrule
  \end{tabular}
    \end{small}
  \end{threeparttable}
\end{table*}

\begin{table*}
  \vskip 0.05in
   \renewcommand{\arraystretch}{1.0} 
  \centering
  \caption{Experimental results of supervised image classification with open-set generative data augmentation on ImageNet-100 {\bf using ViT-B/32 as the classifier.}}\label{tab:appendix_imagenet_100_C_results_vit}
  \begin{threeparttable}
  \begin{small}
  \renewcommand{\multirowsetup}{\centering}
  \setlength{\tabcolsep}{3pt}
  \begin{tabular}{cc|ccc|c|c}
    \toprule
    \multicolumn{2}{c|}{Training Dataset} & 
    \multicolumn{3}{c|}{\rotatebox{0}{\scalebox{1.0}{Data Amount}}} &
    \multirow{2}{*}{\scalebox{1.0}{Real:Syn }} & \scalebox{1.0}{Acc} \\
    \cmidrule(lr){1-2} \cmidrule(lr){3-5} \cmidrule(lr){7-7}    
     \scalebox{1.0}{Real} & \scalebox{1.0}{Syn} & \scalebox{1.0}{Base Real} & \scalebox{1.0}{Added Syn}& Total & & \scalebox{1.0}{Top-1} \\
    \toprule
    \multirow{16}{*}{\scalebox{1.0}{ImageNet-100}}
    & \multirow{16}{*}{\scalebox{1.0}{ImageNet-100-SD3}} & \multirow{5}{*}{\scalebox{1.0}{6,500}} & 0 & 6,500 & real only & \scalebox{1.0}{15.57} \\
    & &  & 650& 7,150 & 1:0.1 & \scalebox{1.0}{15.82 \textcolor{mygreen}{\text{\scriptsize (+0.25)}}} \\
    
    & &  & 3,250 & 9,750 &1:0.5 & \scalebox{1.0}{17.58  \textcolor{mygreen}{\text{\scriptsize (+2.01)}}} \\
     & &  & 6,500 & 13,000 & 1:1& {{\scalebox{1.0}{18.71 \textcolor{mygreen}{\text{\scriptsize (+3.14)}}}}}\\
     & &  & 26,000 & 32,500 & 1:4& {{\scalebox{1.0}{23.25 \textcolor{mygreen}{\text{\scriptsize (+7.68)}}}}} \\
      & &  & 65,000 & 71,500 & 1:10& \scalebox{1.0}{\bf 27.40  \textcolor{mygreen}{\text{\scriptsize (+11.83)}}}  \\

    \cmidrule(lr){3-7}
    & & \multirow{5}{*}{\scalebox{1.0}{26,000}} & 0 & 26,000 & real only & \scalebox{1.0}{27.59}\\
    & &  & 2,600& 28,600 & 1:0.1 & \scalebox{1.0}{28.12 \textcolor{mygreen}{\text{\scriptsize (+0.53)}}}\\
    
    & &  & 13,000 & 39,000 &1:0.5 & \scalebox{1.0}{30.46  \textcolor{mygreen}{\text{\scriptsize (+2.87)}}}  \\
     & &  & 26,000 & 52,000 & 1:1& {{\scalebox{1.0}{32.19 \textcolor{mygreen}{\text{\scriptsize (+4.60)}}}}} \\
      & &  & 104,000 & 130,000 & 1:4& {{\scalebox{1.0}{36.89 \textcolor{mygreen}{\text{\scriptsize (+9.30)}}}}}\\
      & &  & 130,000 & 156,000 & 1:5& \textbf{\scalebox{1.0}{37.82 \textcolor{mygreen}{\text{\scriptsize (+10.23)}}}} \\

    \cmidrule(lr){3-7}
    & & \multirow{4}{*}{\scalebox{1.0}{130,000}} & 0 & 130,000 & real only & \scalebox{1.0}{48.91} \\
    & &  & 13,000& 143,000 & 1:0.1 & \scalebox{1.0}{49.77 \textcolor{mygreen}{\text{\scriptsize (+0.86)}}}  \\
    
    & &  & 65,000 & 195,000 &1:0.5 & \scalebox{1.0}{51.19  \textcolor{mygreen}{\text{\scriptsize (+2.28)}}} \\
     & &  & 130,000 & 260,000 & 1:1& \textbf{{{\scalebox{1.0}{52.31 \textcolor{mygreen}{\text{\scriptsize (+3.40)}}}}}}\\
    
    \bottomrule
  \end{tabular}
    \end{small}
  \end{threeparttable}
\end{table*}

\begin{table*}
  \vskip 0.05in 
  \centering
      \caption{Experimental results of supervised image classification with open-set generative data augmentation on ImageNet-10.}\label{tab:appendix_imagenet_10_AB_results}
  \begin{threeparttable}
  \begin{small}
  \renewcommand{\multirowsetup}{\centering}
  \setlength{\tabcolsep}{7pt}
  \begin{tabular}{cc|ccc|c|c}
    \toprule
    \multicolumn{2}{c|}{Training Dataset} & 
    \multicolumn{3}{c|}{{Data Amount}} &
    \multirow{2}{*}{\scalebox{1.0}{Real:Syn}} &
    \scalebox{1.0}{Acc} \\
    \cmidrule(lr){1-2} \cmidrule(lr){3-5} \cmidrule(lr){7-7}    
     \scalebox{1.0}{Real} & \scalebox{1.0}{Syn} & \scalebox{1.0}{Base Real} & \scalebox{1.0}{Added Syn}& Total & & \scalebox{1.0}{Top-1}\\
    \toprule
    
    \multirow{5}{*}{\scalebox{1.0}{ImageNet-10}}
    & \multirow{5}{*}{\scalebox{1.0}{ImageNet-10-SD2}}& 0 & \multirow{5}{*}{\scalebox{1.0}{13,000}} & 13,000 & syn only & \scalebox{1.0}{59.87} \\
    & & 130 & & 13,130 & 1:100 & \scalebox{1.0}{68.53 \textcolor{mygreen}{\text{\scriptsize (+8.66)}}} \\
    
    & & 1,300 &  & 14,300 &1:10 & \scalebox{1.0}{81.53  \textcolor{mygreen}{\text{\scriptsize (+21.66)}}} \\
     & & 6,500 &  & 19,500 & 1:2& {{\scalebox{1.0}{90.60 \textcolor{mygreen}{\text{\scriptsize (+30.73)}}}}} \\
      & & 13,000 &  & 26,000 & 1:1& \textbf{{\scalebox{1.0}{94.07 \textcolor{mygreen}{\text{\scriptsize (+34.20)}}}}} \\

    \midrule
    \multirow{5}{*}{\scalebox{1.0}{ImageNet-10}}& \multirow{5}{*}{\scalebox{1.0}{ImageNet-10-SD3}} & 
    
    0 & \multirow{5}{*}{\scalebox{1.0}{13,000}} & 13,000 & syn only & \scalebox{1.0}{71.87} \\
    & & 130 & & 13,130 & 1:100 & \scalebox{1.0}{77.20 \textcolor{mygreen}{\text{\scriptsize (+5.33)}}} \\
    
    & & 1,300 &  & 14,300 &1:10 & \scalebox{1.0}{82.80  \textcolor{mygreen}{\text{\scriptsize (+10.93)}}} \\
     & & 6,500 &  & 19,500 & 1:2& {{\scalebox{1.0}{90.80 \textcolor{mygreen}{\text{\scriptsize (+18.93)}}}}} \\
      & & 13,000 &  & 26,000 & 1:1& \textbf{{\scalebox{1.0}{94.13 \textcolor{mygreen}{\text{\scriptsize (+22.26)}}}}} \\
    \bottomrule
  \end{tabular}
    \end{small}
  \end{threeparttable}
  \vspace{5pt}
\end{table*}

\begin{table*}
  \vskip 0.05in
   \renewcommand{\arraystretch}{1.0} 
  \centering
   \caption{Experimental results of zero-shot image classification on ImageNet-10 and ImageNet-100.}\label{tab:appendix_zero-shot_imagenet_ABC_results}
  \begin{threeparttable}
  \begin{small}
  \renewcommand{\multirowsetup}{\centering}
  \setlength{\tabcolsep}{6pt}
  \begin{tabular}{cc|ccc|c|c}
    \toprule
    \multicolumn{2}{c|}{Training Dataset} & 
    \multicolumn{3}{c|}{\rotatebox{0}{\scalebox{1.0}{Data Amount}}} &
    \multirow{2}{*}{\scalebox{1.0}{Real:Syn}} &
    {\scalebox{1.0}{Acc}} \\
    \cmidrule(lr){1-2} \cmidrule(lr){3-5} \cmidrule(lr){7-7}    
     \scalebox{1.0}{Real} & \scalebox{1.0}{Syn} & \scalebox{1.0}{Base Real} & \scalebox{1.0}{Added Syn}& Total & & \scalebox{1.0}{Top-1} \\
    \toprule
    \multirow{5}{*}{\scalebox{1.0}{ImageNet-10}}
    & \multirow{5}{*}{\scalebox{1.0}{ImageNet-10-SD2}} & \multirow{5}{*}{\scalebox{1.0}{6,500}} & 0 & 6,500 & real only & \scalebox{1.0}{45.73} \\
    & &  & 650& 7,150 & 1:0.1 & \scalebox{1.0}{56.27 \textcolor{mygreen}{\text{\scriptsize (+10.54)}}} \\
    & &  & 3,250 & 9,750 &1:0.5 & \scalebox{1.0}{58.93  \textcolor{mygreen}{\text{\scriptsize (+13.20)}}}  \\
     & &  & 6,500 & 13,000 & 1:1& {{\scalebox{1.0}{59.33 \textcolor{mygreen}{\text{\scriptsize (+13.60)}}}}}  \\
      & &  & 65,000 & 71,500 & 1:10& \textbf{{{\scalebox{1.0}{61.33 \textcolor{mygreen}{\text{\scriptsize (+16.60)}}}}}} \\

    \midrule
   \multirow{5}{*}{\scalebox{1.0}{ImageNet-10}}
    & \multirow{5}{*}{\scalebox{1.0}{ImageNet-10-SD3}} & \multirow{5}{*}{\scalebox{1.0}{6,500}} & 0 & 6,500 & real only & \scalebox{1.0}{45.73} \\
    & &  & 650& 7,150 & 1:0.1 & \scalebox{1.0}{62.27 \textcolor{mygreen}{\text{\scriptsize (+16.54)}}}  \\
    
    & &  & 3,250 & 9,750 &1:0.5 & \scalebox{1.0}{67.47  \textcolor{mygreen}{\text{\scriptsize (+21.74)}}} \\
     & &  & 6,500 & 13,000 & 1:1& {{\scalebox{1.0}{68.13 \textcolor{mygreen}{\text{\scriptsize (+22.40)}}}}} \\
      & &  & 65,000 & 71,500 & 1:10& \textbf{{{\scalebox{1.0}{76.80 \textcolor{mygreen}{\text{\scriptsize (+31.07)}}}}}} \\

    \midrule
    \multirow{4}{*}{\scalebox{1.0}{ImageNet-100}}
    & \multirow{4}{*}{\scalebox{1.0}{ImageNet-100-SD3}} & \multirow{4}{*}{\scalebox{1.0}{130,000}} & 0 & 130,000 & real only & \scalebox{1.0}{8.92}\\
    & &  & 13,000& 143,000 & 1:0.1 & \scalebox{1.0}{20.91 \textcolor{mygreen}{\text{\scriptsize (+11.99)}}}  \\
    & &  & 65,000 & 195,000 &1:0.5 & \scalebox{1.0}{34.29  \textcolor{mygreen}{\text{\scriptsize (+25.37)}}} \\
     & &  & 130,000 & 260,000 & 1:1& \textbf{{{\scalebox{1.0}{39.16 \textcolor{mygreen}{\text{\scriptsize (+30.24)}}}}}}  \\
     
    \bottomrule
  \end{tabular}
    \end{small}
  \end{threeparttable}
\end{table*}
\end{document}